\documentclass{article}

\usepackage{microtype}
\usepackage{graphicx}
\usepackage{subfigure}
\usepackage{booktabs} %

\usepackage[breaklinks=true,colorlinks]{hyperref}

\usepackage[accepted]{icml2024}

\usepackage{amsmath}
\usepackage{amssymb}
\usepackage{mathtools}
\usepackage{amsthm}

\def\Secref#1{Section~\ref{#1}}

\def\eqref#1{equation~\ref{#1}}
\def\Eqref#1{Equation~\ref{#1}}

\DeclareMathOperator*{\argmin}{arg\,min}

\usepackage{booktabs} %
\usepackage{multirow} %
\usepackage{graphicx}
\usepackage{tabularx}
\usepackage{xcolor}
\usepackage{amssymb}
\usepackage{setspace}
\usepackage{enumitem}
\usepackage[font={small}]{caption}
\usepackage{wrapfig}
\usepackage{scalerel}
\usepackage{tabu}
\usepackage{titlesec}
\usepackage{float}
\usepackage{makecell}

\usepackage{url}

\definecolor{commentgray}{RGB}{119,119,119}
\newcommand{\numerr}[2]{#1$\pm\mathtt{#2}$}
\newcommand{\commentcode}[1]{\textcolor{commentgray}{\# #1}}
\newcommand{\redt}[1]{\textcolor{black}{#1}}

\newcommand{\redtb}[1]{{\color{black}#1}}
\newcommand{\vspacef}[1]{ \vspace{0pt} } %

\renewcommand{\paragraph}[1]{\noindent\textbf{#1}.}

\usepackage[capitalize,noabbrev]{cleveref}

\theoremstyle{plain}

\theoremstyle{definition}

\theoremstyle{remark}

\usepackage[textsize=tiny]{todonotes}

\icmltitlerunning{Potential Based Diffusion Motion Planning}
\begin{document}

\twocolumn[
\icmltitle{Potential Based Diffusion Motion Planning}

\icmlsetsymbol{equal}{*}

\begin{icmlauthorlist}
\icmlauthor{Yunhao Luo}{brown} %
\icmlauthor{Chen Sun}{brown}
\icmlauthor{Joshua B. Tenenbaum}{mit}
\icmlauthor{Yilun Du}{mit}
\end{icmlauthorlist}

\icmlaffiliation{brown}{Brown University}
\icmlaffiliation{mit}{MIT}
\icmlcorrespondingauthor{Yunhao Luo}{yluo73@cs.brown.edu}

\icmlkeywords{Machine Learning, ICML, Diffusion Model, Motion Planning, Compositionality, Robotics}

\vskip 0.3in
]

\printAffiliationsAndNotice{}  %

\begin{abstract}
Effective motion planning in high dimensional spaces is a long-standing open problem in robotics. One class of traditional motion planning algorithms corresponds to potential-based motion planning. An advantage of potential based motion planning is composability -- different motion constraints can be easily combined by adding corresponding potentials. However, constructing motion paths from potentials requires solving a global optimization across configuration space potential landscape,  which is often prone to local minima. We propose a new approach towards learning potential based motion planning, where we train a neural network to capture and learn an easily optimizable potentials over motion planning trajectories. We illustrate the effectiveness of such approach, significantly outperforming both classical and recent learned motion planning approaches and avoiding issues with local minima. We further illustrate its inherent composability, enabling us to generalize to a multitude of different motion constraints. Project website at \href{https://energy-based-model.github.io/potential-motion-plan}{https://energy-based-model.github.io/potential-motion-plan}.
\end{abstract}

\section{Introduction}

Motion planning is a fundamental problem in robotics and aims to find a smooth, collision free path between a start and goal state given a specified configuration space, and is heavily used across a variety of different robotics tasks such as manipulation or navigation~\citep{laumond1998robot}. A variety of approaches exist for motion planning, ranging from classical 
sampling based approaches~\citep{kavraki1996probabilistic,kuffner2000rrt,karaman2011sampling, gammell2015batch} and optimization based methods~\citep{ratliff2009chomp,mukadam2018continuous,kalakrishnan2011stomp}. A recent body of works have further explored how learned neural networks can be integrated with motion planning for accelerated performance ~\citep{ichter2019robot,qureshi2019motion,fishman2023motion,yamada2023leveraging,le2023accelerating}.

\begin{figure}[t]
  \centering
  \includegraphics[width=1.0\linewidth]{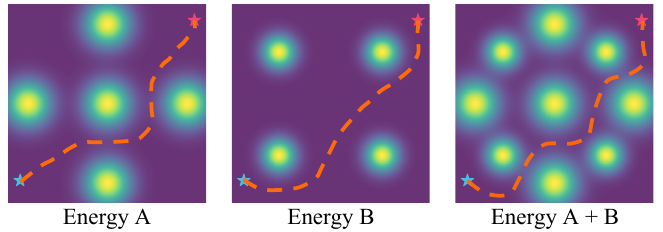}
  \vspace{-15pt} %
  \caption{ \textbf{Illustrative Example of Composing Diffusion Energy Potentials.} Our approach learns different potential functions over motion planning trajectories $q_{1:N}$ (orange dashed lines). Different potentials can be combined and optimized to construct new motion plans that avoid obstacles encoded in both potential functions.}
  \label{fig:teaser}
  \vspace{-15pt}
\end{figure}

A classical approach towards motion planning is potential based motion planning~\citep{khatib1986real,koren1991potential,ratliff2009chomp, ratliff2018riemannian, xie2020geometric}, where both obstacles and goals define energy potentials through which trajectories are optimized to reach. 
An advantage of potential based motion planning is that different constraints to motion planning can be converted into equivalent energy potentials and directly combined to optimize for motion plans. However, potential based methods rely on gradient optimization with respect to local geometry, resulting in the long-standing local minima issues~\citep{lavalle2006planning}. In addition, such motion planning techniques typically require implicit obstacle representations, which is hard to obtain in real-world settings.

We present a potential based motion planning approach leveraging diffusion models~\citep{sohl2015deep,ho2020denoising} where diffusion models are used to parameterize and learn potential landscapes across configuration space trajectories between start and goal states. 
These potential functions can be directly inferred from raw perceptual inputs, removing the need for implicit object representations. Furthermore, the annealed optimization procedure across a sequence of potential energy landscapes in diffusion models~\citep{du2023reduce} can help avoid local minima in optimization.
This optimization procedure is further stochastic, enabling the planner to generate a multitude of motion plans with diverse morphology for a specific problem, offering various motion plan candidates for selection in test time.
Finally, as our potential is defined globally over an environment, it is guided by both local and global environment geometry, and thus our method provides more accurate plans that require significantly less collision checking, compared with problem-independent sampling-based planners.

One major hurdle of learning-based motion planners is their generalizability to environments with unseen, more complex constraints. 
For example, a learned model trained on sparse obstacles will perform poorly in scenarios with cluttered obstacles, as such a setting is out of distribution.
Our approach addresses this through {\it compositionality} -- our learned potentials can be additively composed together to jointly solve motion planning problems with larger sets of constraints than seen at training time.
As illustrated in Figure~\ref{fig:teaser}, combining two potentials from different diffusion models enables us to optimize for trajectories that satisfy both constraints, one to avoid obstacles in a cross, and a second to avoid obstacles in a square. Such flexibility to ad-hoc composition of constraints is especially useful in robotics where agents will often experience new sets of motion constraints in its environment over the course of execution.

Overall, in this paper, our contributions are three-fold. {\textbf{(1)}} We present an approach to learned potential based motion planning using diffusion models. \textbf{(2)} We illustrate the effectiveness of our approach, outperforming existing classical and learned motion planning algorithms in accuracy and collision checks.
\textbf{(3)} We illustrate the compositionality of motion planner, enabling it to generalize to multiple sets of motion constraints as well as an increased number of objects.

\section{Related Work}
\textbf{Motion Planning.} 
Classic sampling-based motion planners~\citep{kavraki1996probabilistic, kuffner2000rrt, elbanhawi2014sampling, gammell2014informed, janson2015fast, choudhury2016regionally, strub2020advanced} have gained wide adoption due to their efficiency and generalizability.
However, problem-independent nature of these methods can result in inefficiency particularly when planning for similar problems repetitively.
Reactive methods, such as potential-based approaches~\citep{khatib1986real, ratliff2018riemannian, xie2020geometric}, velocity obstacles~\citep{fiorini1998motion, van2008reciprocal}, and safety barrier certificates~\citep{wang2017safety} can provide fast updates and have the guarantee for obstacle avoidance.
However, their performance is typically constrained by local minima or numerical instability~\citep{lavalle2006planning}, and they usually need to construct obstacle representations in the robot configuration space, which is hard to obtain from raw perception.
To address these issues, recent works have proposed many deep-learning based motion planners~\citep{ichter2019robot, bency2019neural, fishman2023motion}. %
Other works combine neural network with sampling-based methods~\citep{qureshi2019motion, johnson2021motion, yu2021reducing, lawson2022control}, or combine trajectory level generative models with  specifically designed cost functions~\citep{saha2023edmp, carvalho2023motion} that requires privileged information.
These methods can increase planning speed, expand the planning horizon, or reduce the access queries to the environment by leveraging learned knowledge.
However, the performance of learned motion-planning approaches on out-of-distribution environments often sharply degenerates.
In addition, many existing methods are only constrained to simple 2D environments \citep{yonetani2021path,chaplot2021differentiable,toma2021waypoint,chang2023denoising,carvalho2022conditioned}.
In contrast, we present a learning-based potential motion planning approach, requiring no ground truth knowledge of the optimized cost objective, which we illustrate can effectively generalize to new environments through composability of potentials.

\looseness=-1
\textbf{Diffusion Models for Robotics.} 
Many recent works have explored the application of diffusion model for robotics~\citep{janner2022diffuser, chen2022offline, kapelyukh2023dall, ha2023scaling}. 
Current research spans a variety of robotics problems, including  action sequence generation~\citep{liang2023adaptdiffuser, fang2023dimsam, li2023hierarchical}, policy~\citep{wang2023diffusion, kang2023efficient},
grasping~\citep{urain2023se, Huang_2023_CVPR}, 
and visuomotor planning or control \citep{dalal2023imitating, yang2023probabilistic, chi2023diffusion}, with recent work also exploring their application in solving manipulation constraints~\citep{yang2023compositional}.
In contrast to these works, we focus on how diffusion models can be used to explicitly parameterize and learn potentials in potential-based motion planning. 
We illustrate the efficacy of such an approach across motion-planning settings and its ability to generalize to new environments through composition with other learned potentials.

\begin{figure*}[t]
  \centering
  \includegraphics[width=1.0\linewidth]{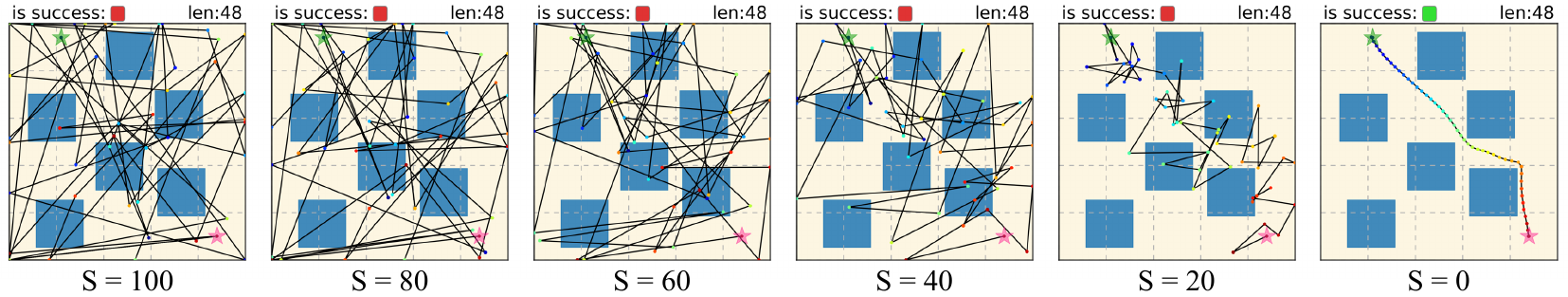}
\caption{ \textbf{Trajectory Denoising Process.} The trajectory is randomly initialized from Gaussian in timestep $S = 100$. Noise is iteratively removed via the gradient of the energy function as given in \eqref{eqn:diffusion_opt}. A feasible trajectory can be obtained at timestep $S = 0$.}
  \label{fig:22-energy-per-step}
  \vspacef{-25pt}
\end{figure*}

\section{Method}
We first introduce potential based motion planning in \Secref{sect:potential}. We then discuss how potential based motion planning can be implemented with diffusion models in \Secref{sect:diffusion_potential}. We further discuss how such an approach enables us to combine multiple different potentials together in \Secref{sect:diffusion_composition}. Following this, we discuss how we can refine motion plans generated by diffusion models in cases of collision in \Secref{sect:motion_refine}. Finally, we show the probabilistic completeness of the proposed method in \Secref{sect:completeness}.

\vspacef{-10pt}
\subsection{Potential Based Motion Planning}
\label{sect:potential}
Given a specified start state $q_{\text{st}}$ and end state $q_{\text{e}}$ in a configuration space $\mathbb{R}^n$, motion planning is formulated as finding a collision-free trajectory $q_{1:N}$ which starts from $q_{\text{st}}$ and ends at $q_{\text{e}}$. 
To solve for such a collision-free trajectory $q_{1:N}$ in potential based motion planning~\citep{khatib1986real,koren1991potential}, a potential function $U(q): \mathbb{R}^n \rightarrow \mathbb{R}$ on the configuration space composed of

\vspace{-12pt}
\begin{equation}
    U(q) = U_{\text{att}}(q) + U_{\text{repel}}(q)
    \label{eqn:potential}
\end{equation}
is defined, where \redt{$U(q)$} assigns low potential value to the goal state $q_{\text{e}}$ and high potential to all states which are in collision. In \Eqref{eqn:potential}, $U_{\text{att}}(q)$ represents a attraction potential that has low values at the the end state $q_{\text{end}}$ and high values away from it, and $U_{\text{repel}}(q)$ represents a repulsion potential that has high values near obstacles and low values away them. The functional form of the potential function $U(q)$ provides an easy approach to integrate additional obstacles in motion planning by adding \redt{a} new potential $U_{\text{new}}(q)$ representing obstacles to the existing potential in \Eqref{eqn:potential}.

\redt{To obtain a motion plan from a potential function $U(q)$, a collision-free trajectory $q_{1:N}$ from $q_{\text{st}}$ to $q_{\text{e}}$ is obtained by iteratively following the scaled gradient of the potential function}
\begin{equation}
    q_t = q_{t-1} -\gamma \nabla_q U(q)
    \label{eqn:opt}
\end{equation}
with a successful motion plan constructed when the optimization procedure reaches the minimum of the potential function $U(q)$. A major limitation of above approach in \Eqref{eqn:opt} is {\it local minima} -- if the optimization procedure falls in such a minima, the motion plan will no longer successfully construct paths from $q_{\text{st}}$ to $q_{\text{e}}$~\citep{yun1997wall,teli2021fuzzy,lavalle2006planning}.

\subsection{Potential Based Diffusion Motion Planning}
\label{sect:diffusion_potential}

We next discuss how to learn potentials for potential motion planning that enables us to effectively optimize samples. Given a motion plan $q_{1:T}$ from start state $q_{\text{st}}$ to end state $q_{\text{e}}$ and a configuration space characterization $C$ ({\it i.e.} the set of obstacles in the environment), we propose to learn a trajectory-level potential function $U_\theta$ so that
\begin{equation}
    q_{1:T}^* = \argmin_{q_{1:T}} U_\theta(q_{1:T}, q_{\text{st}}, q_{\text{e}}, C)
    \label{eqn:potential_traj}
\end{equation} 
where $q_{1:T}^*$ is a \redt{successful} motion plan from $q_{\text{st}}$ to $q_{\text{e}}$. 
To learn the potential function in \Eqref{eqn:potential_traj}, we propose to learn an EBM~\citep{lecun2006tutorial,du2019implicit} across a dataset of solved motion planning $D = \{q_{\text{st}}^i, q_{\text{e}}^i,  q_{1:T}^i, C^i \}$, where $e^{-E_\theta(q_{1:T}|q_{\text{st}}, q_{\text{e}}, C)} \propto p(q_{1:T}|q_{\text{st}}, q_{\text{e}}, C)$. Since the dataset $D$ is of solved motion planning problems, the learned energy function $E_\theta$ will have minimal energy at successful motion plans $q_{1:T}^*$ and thus satisfy our potential function $U_\theta$ in \Eqref{eqn:potential_traj}.

To learn the EBM landscape that enables us to effectively optimize and generate motion plans $q_{1:T}^*$, we propose to shape the energy landscape using denoising diffusion training objective~\citep{sohl2015deep,ho2020denoising}. In this objective, we explicitly train the energy landscape so gradient with respect to the energy function can denoise and recover a motion plan $q_{1:T}$ across many differing levels of noise corruption $\{1, \ldots, S\}$ ranging from mostly correct motion paths to fully corrupted Gaussian noise trajectories. By shaping the gradient of the energy function to generate motion plans $q_{1:T}$ from arbitrary \redt{initialized trajectories}, our learned energy landscape is able to effectively optimize for motion paths.

Formally, to train our potential, we use the energy based diffusion training objective in ~\citep{du2023reduce}, 
\redt{where the gradient of energy function is trained to denoise noise-corrupted motion plans $q_{1:T}^i$ from $D$,}
\begin{equation}
\scalebox{0.85}{ $
 \mathcal{L}_{\text{MSE}}=\|\mathbf{\epsilon} - \nabla_{q_{1:T}} E_\theta(\sqrt{1-\beta_s} q_{1:T}^i +  \sqrt{\beta_s} \mathbf{\epsilon}, s, q_{\text{st}}^i, q_{\text{e}}^i, C^i)\|^2
 $}
 \label{eqn:train_obj}
\end{equation}
where $\epsilon$ is sampled from Gaussian noise $\mathcal{N}(0, 1)$,  $s \in\{1,2,...,S\}$ is the denoising diffusion step \redt{(we set $S = 100$)}, and $\beta_s$ is the corresponding Gaussian noise corruption on a motion planning path $q_{1:T}^i$. We refer to $E_\theta$ as the {\it diffusion potential function}.

To optimize and sample from our diffusion potential function, we initialize a motion path $q_{1:T}^S$ at diffusion step $S$ from Gaussian noise $\mathcal{N}(0, 1)$ and optimize for motion path following the gradient of the energy function \redt{$E_{\theta}$}. We iteratively refine the motion path $q_{1:T}^s$ across each diffusion step following 
\vspacef{-5pt}
\begin{align}
& q_{1:T}^{s-1}=q_{1:T}^{s}-\gamma \epsilon + \xi, \quad \xi \sim \mathcal{N} \bigl(0, \sigma^2_s I \bigl), \notag \\
& \epsilon =  \nabla_{q_{1:T}} E_\theta(q_{1:T}, s, q_{\text{st}}, q_{\text{e}}, C)
\label{eqn:diffusion_opt}
\end{align}
\vspace{-19pt}

To parameterize the energy function $E_\theta(q_{1:T}, s, q_{\text{st}}, q_{\text{e}}, C)$, we use classifier-free guidance scale of 2.0~\cite{ho2022classifier} to form a peaker composite energy function conditioned on $C$.
$\gamma$ and $\sigma^2_s$ are diffusion specific scaling constants\footnote{A rescaling term at each diffusion step is omitted above for clarity}. The final predicted motion path $q_{1:T}^*$ corresponds to  the output $q_{1:T}^0$ after running $S$ steps of optimization from the diffusion potential function.

\begin{figure}[t]
\vspace{-10pt}
\centering
\small
\begin{minipage}{\linewidth}
\begin{algorithm}[H]
\caption{Compositional Potential Based Planning}
\label{alg:comb}
\begin{algorithmic}[1]
  \STATE \textbf{Models:} compositional set of $N$ diffusion potential functions $E^i_\theta( q_{1:T}, t, q_{\text{st}}, q_{\text{e}}, C_i )$
  \STATE \textbf{Hyperparameters:} trajectory horizon $T$, number of denoising diffusion steps $S$ 
  \STATE \textbf{Input:} start position $q_{\text{st}}$, end position $q_{\text{e}}$, $N$ motion planning constraints $C_{1:N}$
  
  \STATE Initialize $ q_{1:T}^S \sim \mathcal{N}(0, I) $
  \FOR{ $s$ = $S \dots 1$ }
    \STATE \commentcode{Combining Different Energy Potentials Together}
    
      \STATE $ \epsilon_{\text{comb}} = \sum_{i=1}^{N} \nabla_{q_{1:T}} E_\theta^{i}(q_{1:T}^s, s, q_{\text{st}}, q_{\text{e}}, C_i)  $

    \STATE \commentcode{Transit to Next Diffusion Time Step}
    \STATE $q_{1:T}^{s-1} = q_{1:T}^{s}-\gamma \epsilon_{\text{comb}}  + \xi, \quad \xi \sim \mathcal{N} \bigl(0, \sigma^2_s I \bigl)$

  \ENDFOR
  \STATE \textbf{return} $q_{1:T}^0$
\end{algorithmic}
\end{algorithm}
\end{minipage}
\vspace{-5pt}
\end{figure}

\vspacef{-15pt}
\subsection{Composing Diffusion Potential Functions}
\label{sect:diffusion_composition}
\vspacef{-4pt}
Given two separate diffusion potential functions $E^1_\theta(\cdot)$ and $E^2_\theta(\cdot)$, encoding separate constraints for motion planning, we can likewise form a composite potential function $E^{\text{comb}}(\cdot) = E^1(\cdot) + E^2(\cdot)$ by directly summing the corresponding potentials. \redt{This potential function $E^{\text{comb}}$ will have low energy precisely at motion planning paths $q_{1:T}$ which satisfy both constraints, with sampling corresponding to optimizing this potential function.}
To sample from the new diffusion potential function $E^{\text{comb}}$, we can follow

\vspace{-5pt}
\scalebox{0.93}{

\begin{minipage}{\linewidth}
\begin{align}
    &q_{1:T}^{s-1}=q_{1:T}^{s}-\gamma \epsilon^{\text{comb}}+\xi, \quad \xi \sim \mathcal{N} \bigl(0, \sigma^2_s I \bigl), \label{eqn:diffusion_opt_comb} \\
    &\epsilon^{\text{comb}} = \nabla_{q_{1:T}} (E_\theta^{1}(q_{1:T}, s, q_{\text{st}}, q_{\text{e}}, C_1) + E_\theta^{2}(q_{1:T}, s, q_{\text{st}}, q_{\text{e}}, C_2)) \notag
\end{align}
\end{minipage}
}
\vspace{3pt}

This composite potential is the ground truth potential combining constraints if the constraints are independent, which is satisfied if the set of trajectories given constraints are uniformly distributed (see Appendix \ref{sect:app_cond_indep}), and otherwise serves as approximate proxy to optimize multiple constraints.

\textbf{Applications of Composing Potential Functions.} The ability to combine multiple separate potential functions for motion planning offers a variety of different ways to generalize and extend existing motion planning systems. First, in many motion planning problems, there is often a heterogenous set of different types of constraints or collisions that limit possible configuration space paths. For instance, in autonomous driving, constraints that can arise may include moving pedestrians, traffic lanes, road work, or incoming cars. Oftentimes, we cannot enumerate all potential combinations, but we wish motion planning systems to be able to handle all possible combinations of constraints. Jointly learning a single motion planning model for all constraints may be difficult, as at test time, we may see novel combinations that we do not have training data for. By learning separate diffusion potential fields for each constraint, we can combine them in an ad-hoc manner at test time to deal with arbitrary sets of constraints. We provide two concrete implementations of composing potentials together below and a detailed procedural in Algorithm \ref{alg:comb}.

\paragraph{Generalization over More Obstacles} Suppose that the \redt{potential function $E_\theta$} is trained on environments with 4 obstacles $|C| = 4$. However, in test time, we want to generalize to a more complex environment that has 6 obstacles $C^\prime = \{o_1, o_2, o_3, o_4, o_5, o_6\}$.
This can be achieved by adding the potentials evaluated on two sets of obstacles, where $C_1 = \{o_1, o_2, o_3, o_4 \}$ and $C_2 = \{o_3, o_4, o_5, o_6 \}$.
This formulation can be further extended to $N$ sets of obstacles $C_{1:N}$ and the composite diffusion potential function is given by:

\vspace{-17pt}
\begin{equation}
\scalebox{0.95}{$
E_\theta^{\text{comb}}(q_{1:T}, s, q_{\text{st}}, q_{\text{e}}, C_{1:N} )
=  \sum\limits^{N}_{i=1} E_\theta(  q_{1:T}, s, q_{\text{st}}, q_{\text{e}}, C_i  )
$}
\label{eq:moreC_comb}
\end{equation}

\vspace{-5pt}

\begin{figure}[t]
  \centering
  \includegraphics[width=1.0\linewidth]{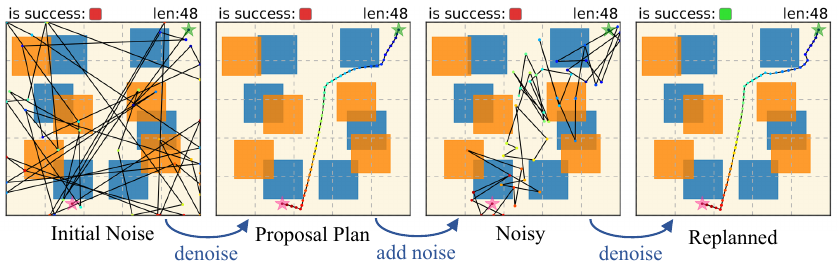}
  \vspace{-12pt}
  \caption{ \textbf{Visualization of the Motion Refining Scheme}. A proposal plan is first generated by denoising an initial Gaussian noise. If collision is detected, a small noise is added to the proposal and the new plan is generated by denoising the partially noisy trajectory.}
  \label{fig:replan}
  \vspace{-10pt}
\end{figure}

\textbf{Generalization over Static and Dynamic Obstacles.} 
Many real-life scenarios involve \textit{dynamic} real-time interaction. For instance, to construct motion plans for an autonomous vehicle, we must both avoid static lane obstacles as well as moving cars. %
\redt{While static obstacles are often known a priori, the motion patterns of dynamics obstacles often change with time, making it advantageous to be able to combine different dynamic constraints with static ones.}
We can directly implement this by adding diffusion potential function $E_{\theta_s}^{i}$ that only trained on static obstacles 
$C^s_i$ and diffusion potential function $E_{\theta_d}^{j}$ that only trained on dynamic obstacles $C^d_j$. 
In a more general form,
to condition on
a set of $N_1$ static obstacles $C^s_{1:N_1}$ with their diffusion potential functions $E_{\theta_s}^{1:N_1}$ and a set of $N_2$ dynamic  $C^d_{1:N_2}$ obstacles with their diffusion potential functions $E_{\theta_d}^{1:N_2}$,
the composite diffusion potential function can be written as:

\vspace{-15pt}
\begin{equation}
\small
\label{eq:dynC_comb}
\scalebox{1.0}{$
\begin{aligned}
 & E_{\theta}^{\text{comb}}(q_{1:T}, s, q_{\text{st}}, q_{\text{e}}, [C_{1:N_1}^s, C_{1:N_2}^d] ) = \\
 & \sum^{N_1}_{i=1} E_{\theta_s}^{i}( q_{1:T}, s, q_{\text{st}}, q_{\text{e}}, C_i^s) + \sum^{N_2}_{j=1} E_{\theta_d}^{j}( q_{1:T}, t, q_{\text{st}}, q_{\text{e}}, C_j^d)
\end{aligned}
$}
\end{equation}
\vspacef{-15pt}
\begin{algorithm}[t]
\caption{Refining Motion Plans}
\label{alg:replan}
\begin{algorithmic}[1]
  \STATE \textbf{Model:} compositional potential denoiser  
  $f_\theta( q_{1:T}, s, q_{\text{st}}, q_{\text{e}}, C_{1:N} )$
  
  \STATE \textbf{Hyperparameters:} number of refine attempts $R$, noise scale $k$
  \STATE \textbf{Input:} trajectory $q_{1:T}$, start position $q_{\text{st}}$, end position $q_{\text{e}}$, $N$ constraints $C_{1:N}$
  
  \STATE $Z$ = $\mathtt{Get\_Collision\_Sections(}$$q$\texttt{)} \ \hfill \comment{A Set of Indices of Collision Sections in $q_{1:T}$}
  \FOR{ $r$ = $1 \dots R$ }
    \STATE $q_{1:T}^k = \sqrt{ \bar{\alpha}_k } q_{1:T} + (1 - \bar{\alpha}_k) \xi,  \quad \xi \sim \mathcal{N} \bigl(0, \sigma^2_t I \bigl)$ \ \hfill \comment{Add Noise to $q_{1:T}$}
    \STATE $ q^{\prime} = f_\theta(q_{1:T}^{k}, k,  q_{\text{st}}, q_{\text{e}}, C_{1:N})$, \hfill \comment{Get new Denoised Trajectory}
      \FORALL{  $z_i \in Z$ }
            \IF { $\mathtt{is\_section\_good}$($ q^{\prime}[z_i] $)}
                \STATE $q[z_i] = q^{\prime}[z_i] $; \; $Z = Z \setminus z_i $ \ \hfill \comment{Refine $q_{1:T}$ and Remove $s_i$ from set $S$}
            \ENDIF
        
      \ENDFOR      
  \ENDFOR
\STATE \textbf{return} $q_{1:T}$
\end{algorithmic}
\end{algorithm}

\vspace{-10pt}
\subsection{Refining Motion Plans}
\label{sect:motion_refine}
In practice, the predicted motion plan $q_{1:T}$ might occasionally contains sections that violate the constraints of the environment (i.e., collide with obstacles). 
To tackle this issue, both classical and learned motion planners~\citep{kuffner2000rrt, qureshi2019motion}  provide mechanisms to refine trajectories subject to collisions in configuration space. %
With diffusion potential fields, we can likewise refine a trajectory,  $q_{1:T}$ with collision, by locally perturbing it into a noisy trajectory $q_{1:T}^k$ defined by the $k$-th step of the diffusion forward process:
\vspacef{-5pt}
\begin{equation}
    q_{1:T}^k = \sqrt{ \bar{\alpha}_k } q_{1:T} + (1 - \bar{\alpha}_k) \xi,  \quad \xi \sim \mathcal{N} \bigl(0, \sigma^2_t I \bigl)
    \vspacef{-5pt}
\end{equation}
\redt{as in \cite{ho2020denoising}.} A new motion plan $q_{1:T}^\prime$ can be obtained by denoising the noisy trajectory following Equation \ref{eqn:diffusion_opt}, 
where $q_{1:T}^\prime = f_\theta( q_{1:T}^k, k, q_{\text{st}}, q_{\text{e}}, C_{1:N} )$ and  $f_\theta(\cdot)$ is an iterative diffusion potential denoiser that outputs a clean trajectory. 
To fix the problematic trajectory $q_{1:T}$, the collision sections in $q_{1:T}$ will be replaced by the corresponding sections in $q_{1:T}^\prime$ if these new sections are coherent and collision-free. This refining procedural can be repeated until a desired trajectory is found.
The warm-start denoising scheme can enable faster re-planning and maintain the morphology of the original plan, supporting planning on energy-critical mobile devices or when the plan has been executed. 
Algorithm \ref{alg:replan} displays the complete refining pipeline and Figure \ref{fig:replan} provides a corresponding visualization.

\vspacef{-7pt}
\subsection{Probabilistic Completeness}\label{sect:completeness}
In this section, we will show the probabilistic completeness of our method.
Let $f_\theta(q_{1:T})$ 
denote the probability density function of the output distribution $\mathcal{D}_o$ of our diffusion potential model.
In such learned neural distribution, all data points $q_{1:T}$ are assigned positive density, that is, 
\vspacef{-5pt}
\begin{equation}
    \forall q_{1:T}, \ f_\theta( q_{1:T} ) > 0
    \label{eqn:positive_density}
\end{equation}
\vspace{-20pt}

Define $\mathcal{J}_c$ as the a set of all valid trajectories subject to constraint $C$.
There always exists a small interval in the vicinity of a random trajectory $q_{1:T}^c \in \mathcal{J}_c$, such that
\begin{equation}
    [q_{1:T}^c - \tau, q_{1:T}^c + \tau] \subseteq	 \mathcal{J}_c, \quad \tau > 0
\end{equation}
i.e., all trajectories in the interval satisfy the given constraint $C$.
Let $\mathbb{P}_\tau$ denote the probability for our model to
sample a trajectory from the interval,
and according to \eqref{eqn:positive_density} we have,
\begin{equation}
    \mathbb{P}_{\tau} = \int_{q_{1:T}^c - \tau}^{q_{1:T}^c + \tau} f(x) dx  > 0
\end{equation}
Let $A_n$ denote the event that there is at least one trajectory $q_{1:T} \in \mathcal{J}_c$ among $n$ sampled trajectories.
Clearly, as the number of samples approaches infinity, event $A$ will happen almost surely, i.e.,
\vspace{-5pt} %
\begin{equation}
    \lim_{n \to \infty} \mathbb{P} \ \big( A_n \big) = 1
    \vspace{-2pt} %
\end{equation}
Hence, our method is probabilistically complete.

\begin{figure}[t]
  \centering
  \includegraphics[width=1.0\linewidth]{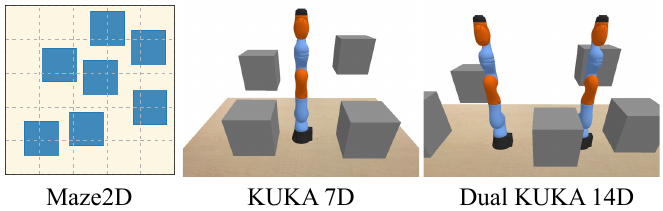}
  \vspace{-14pt}
  \caption{ \textbf{Environment Demonstration.} Maze2D: a point robot moving in 2D workspace with the highlighted blocks as obstacles. KUKA: robotic arm with 7 DoF operating on a tabletop. The grey cuboids are obstacles. Dual KUKA 14D: Two side by side KUKA arms operate simultaneously.}
  \label{fig:2-env_show}
  \vspace{-10pt}
\end{figure}

\begin{figure*}[t]
  \centering
  \includegraphics[width=1.0\linewidth]{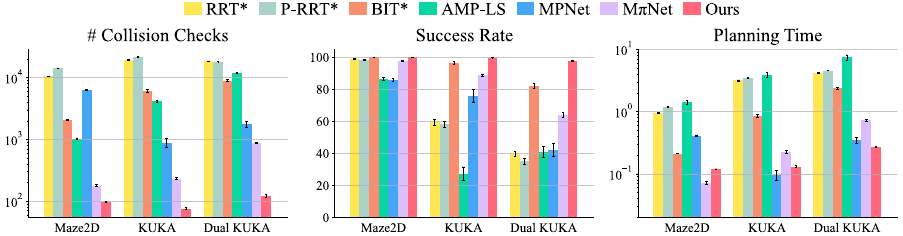}
  \vspace{-17pt}
  \caption{ 
  \textbf{Quantitative Comparisons in Motion Planning Environments.} Three metrics of three environments from 2D to 14D are reported. From left to right: a) number of collision checks, b) success rate, c) planning time.}
  \label{fig:base_barplot}
  \vspace{-5pt}
\end{figure*}

\section{Experiments}

\label{sec:exp}
In this section, we first describe our environments and baselines in Section \ref{subsec:env}.
Next, in Section \ref{subsec:base}, we discuss our experiments on the base environments and the motion refining algorithm.
Following, in Section \ref{subsec:comp}, we present the compositionality results by evaluating our motion planner on composite environments. Then, we describe the real world motion planning performance in Section \ref{subsec:realworld}. 

\subsection{Environments and Baselines} \label{subsec:env}
We first classify the environments that we evaluated on to 3 categories by the level of generalization capability:

\vspace{-5pt}

\begin{itemize}[leftmargin=*, itemsep=-0.0em]
    \item \textbf{Base Environment:} same number of constraints as in training phase, constraints sampled from the same distribution as in training.
    \item \textbf{Composite Same Environment:} more constraints than training phase, constraints sampled from the same distribution as in training.
    \item \textbf{Composite Different Environment:} more constraints than training phase, constraints sampled from different distributions.
\end{itemize}
\vspace{-3pt}

\vspace{-5pt}
Our simulated motion planning environments are listed below and shown in Figure \ref{fig:2-env_show}. See Appendix \ref{sect:app_env_dataset} for more details. In all environments, the motion planning task is to generate a feasible trajectory from the start state to the goal state. The agent trajectory is represented in C-space, while obstacles are represented in workspace.

\vspace{-5pt}

\begin{itemize}[leftmargin=*, itemsep=-0.0em]
\item 
\textbf{Maze2D}. A point-robot moving in a 2D workspace. The configuration space is the x-y coordinate of the agent. We offer two variants: \textit{Static Maze2D} where obstacles stay in the same locations and \textit{Dynamic Maze2D} where obstacles are moving in randomly generated linear trajectories. 
\item 
\textbf{KUKA}. A KUKA arm of 7 DoF operating on a tabletop in a 3D workspace. The start/goal is given as the 7D joint state of the KUKA arm. 
\item 
\textbf{Dual KUKA}. Two KUKA arms are placed side by side on a tabletop and operating simultaneously. %
The start/goal is given as the joint states of two KUKA arms (14 DoF). 
\end{itemize}

\vspace{-2pt}

\begin{figure}[t]
  \centering
  \includegraphics[width=1.0\linewidth]{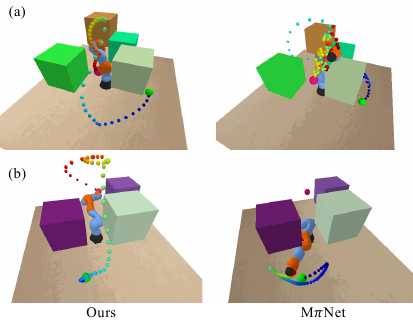}
  \vspace{-16pt}
  \caption{\textbf{Qualitative Motion Plans in KUKA Environment.} 
  Results of two motion planning problem are shown in two rows. The large green/pink ball indicates the start/goal state. Our method in the left column generates smooth and near-optimal trajectories, while in row (a), the trajectory of M$\pi$Net chooses a longer route from behind and gets stuck in some local regions, and in row (b), M$\pi$Net cannot pass through the narrow passage and keeps hovering near the start state.
  }
  \vspace{-13pt}
  \label{fig:8-kuka_base}
\end{figure}

\paragraph{Baselines} For non-learning motion planning methods, our baselines include classic sampling-based planning baseline RRT*~\citep{karaman2011sampling}, sampling-based method with potential functions based heuristic {P-RRT*~\citep{qureshi2016potential}, advanced sampling-based method BIT*~\citep{gammell2015batch}}, sampling-based method for dynamic environments SIPP~\citep{phillips2011sipp}, and {traditional potential-based method RMP~\citep{ratliff2018riemannian}}.
For learned motion planners, we compare our method with: MPNet~\citep{qureshi2019motion}, M$\pi$Net~\citep{fishman2023motion}, and AMP-LS~\citep{yamada2023leveraging}. 
MPNet is trained on trajectories with sparse waypoints and use MLPs to encode environment configuration and predict the next robot state. 
M$\pi$Net is trained on dense trajectory waypoints and predicts the movement vector instead of the next robot state.
AMP-LS encodes the robot state into a latent feature and approaches the goal state by iteratively using the gradient of several planning losses to update the latent. A sequence of latents are then decoded and form a trajectory.
In evaluation, all the start/goal states and environment configurations are unseen to the models. For each experiment, we evaluate on 100 different environments with 20 randomly sampled starts and goals in each environment. No re-training is performed in test time.

\subsection{Performance on Base Environments}\label{subsec:base}
We first evaluate motion planning performance in each base environment: randomly generated environments that follow the same procedural generation pipeline as the training environments. Quantitative results are shown in Figure \ref{fig:base_barplot} 
and Appendix \ref{sect:app_base_env}.
We include the full details of evaluation setup in Appendix \ref{app:eval-details}.

\paragraph{Comparison to Sampling-based Planners}
We set the planning time limit to 5 seconds for all sampling-based methods. Both RRT*, P-RRT*, and BIT* can succeed in the base Maze2D environment. However, their success rates suffer from a significant degradation when the dimension of the configuration space increases. 
The planning time of the sampling-based planners also rises dramatically as the dimension of the space increases.
\redt{By contrast, the success rates of our method surpass all baselines and are consistent across different environments, achieving this in less than 11\% of BIT*'s planning time and requiring up to two orders of magnitude fewer collision checks.}

\begin{figure}[t]
  \centering
  \includegraphics[width=1.0\linewidth]{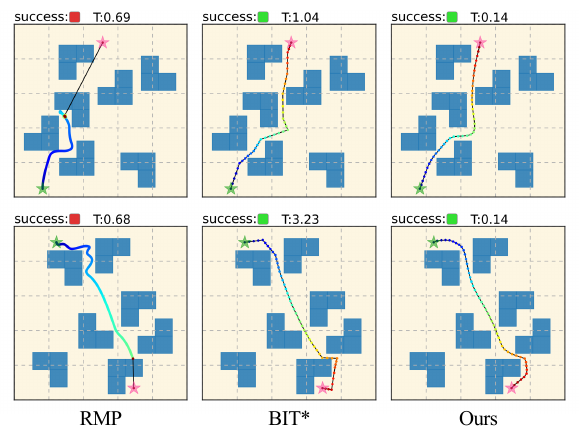}
  \vspace{-13pt}
  \caption{
  \textbf{Qualitative Performance on Environments with Concave Obstacles.} 
  Potential-based planning method RMP is prone to local minima. BIT* is able to find a feasible solution, but the trajectories are rough and require much longer planning time, especially for the harder problem in the second row. Trajectories generated by our planner are both smooth and short in length.
  }
  \label{fig:21-concave7_compare-main-paper}
  \vspace{-15pt}
\end{figure}

\begin{table}[b]
\vspace{-18pt}
\setlength{\tabcolsep}{4pt}
\centering
\small %
\begin{tabular}{ c c c c c c c}

 & \multicolumn{2}{ c }{$R = 3$} & \multicolumn{2}{ c }{$R = 5$} & \multicolumn{2}{ c }{$R = 10$} \\
 \cmidrule(lr){2-3} \cmidrule(lr){4-5} \cmidrule(lr){6-7}
Env  &  Before & After &  Before & After &  Before & After \\
\midrule
Maze2D & 96.3  & 99.8  & 95.3 & 99.0 & 95.8 & 100.0 \\
KUKA &  71.3 &   90.0   & 69.5 & 94.3 & 69.8 & 94.8 \\
Dual KUKA &  45.5 & 69.8 & 47.3 & 77.3 & 47.0 & 80.8 \\
\bottomrule
\end{tabular}
\vspace{-5pt}
\captionof{table}{\textbf{Quantitative Results of Motion Plan Refining}. Success rates before and after motion refining are shown. $R$ denotes the number of refining attempts. The proposed refining method can consistently boost success rates on three base environments.}
\label{tab:replan}
\end{table}

\paragraph{Comparison to Learning-based Planners}
We also compare to three learning-based motion planning baselines: MPNet, M$\pi$Net, and AMP-LS.
Our method outperforms all baselines in both success rate and number of collision check.
In Dual KUKA, our method leads the state-of-the-art learning-based planner M$\pi$Net by nearly 35\% in success rate while with less than 40\% of its planning time and 7 times less of its collision checks.
\redt{We also observe that the performance of other planners drop quickly as the difficulty increases, while our planner performs steadily across all environments. Though the planning time of M$\pi$Net in Maze2D is slightly shorter than ours, the gap is closing as the dimension of the configuration space increases, and in Dual KUKA, our planner requires less planning time than all baselines, probably due to the expensive cost of collision checks and higher task complexity. Qualitative comparisons are shown in Figure \ref{fig:8-kuka_base} and \ref{fig:41-kuka7d-base-app}.
}

\begin{figure}[t]
  \centering
  \includegraphics[width=1.0\linewidth]{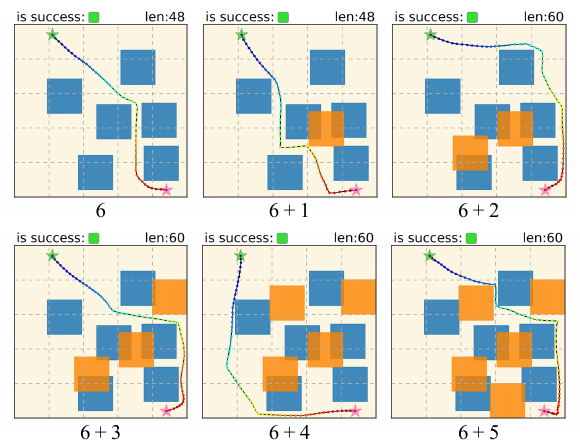}
  \vspace{-15pt}
  \caption{  \textbf{Qualitative Compositional Generalization over More Obstacles.} Two models that trained on only 6 obstacles are composed and tested on out-of-distribution environments with 7, 8, 9, 10, 11 obstacles, respectively.}
  \vspace{-4pt}
  \label{fig:6-comp_rm2d}
\end{figure}

\begin{table}[t]
\small\setlength{\tabcolsep}{5.5pt}
\centering
\small %
\begin{tabular}{ccccc}
 & \multicolumn{2}{ c }{Maze2D -- Convex} & \multicolumn{2}{ c }{Maze2D -- Concave} \\
\cmidrule(lr){2-3} \cmidrule(lr){4-5} %

Method & Success & Time & Success & Time \\
\midrule
RRT*  & 98.8 & 0.95  & 92.1 & 2.14 \\ %
P-RRT* & 98.5 & 1.17 & 85.9 & 2.69 \\ %
BIT* & 100.0 & 0.21 & 100.0 & 0.45 \\ %
MPNet  & 88.4 & 0.21 & 84.3  & 0.38  \\
M$\pi$Net & 97.9 & \textbf{0.07} & 96.9 & \textbf{0.10}  \\ %
MPD & 77.9  & 2.99 &  44.4 & 3.93 \\ 
RMP &   64.9  & 0.13  &  28.0 & 0.34 \\
Ours  & \textbf{100.0} & \textbf{0.12} & \textbf{100.0} & \textbf{0.15}\\ %

\bottomrule
\end{tabular}
\caption{ %
\textbf{Quantitative Performance on Convex and Concave Obstacles.}
Motion planning performance on Maze2D environments with 6 convex obstacles (left) and 7 concave obstacles (right).
The proposed diffusion potential planner outperforms the traditional potential-based method RMP and diffusion motion planning counterpart MPD by a margin. While both our method and the advanced sampling-based method BIT* can solve all the problems, ours requires significantly less planning time than BIT*.
}
\vspace{-10pt}
\label{tab:concave7-main}
\end{table}

\begin{figure}[t]
  \centering
  \includegraphics[width=1.0\linewidth]{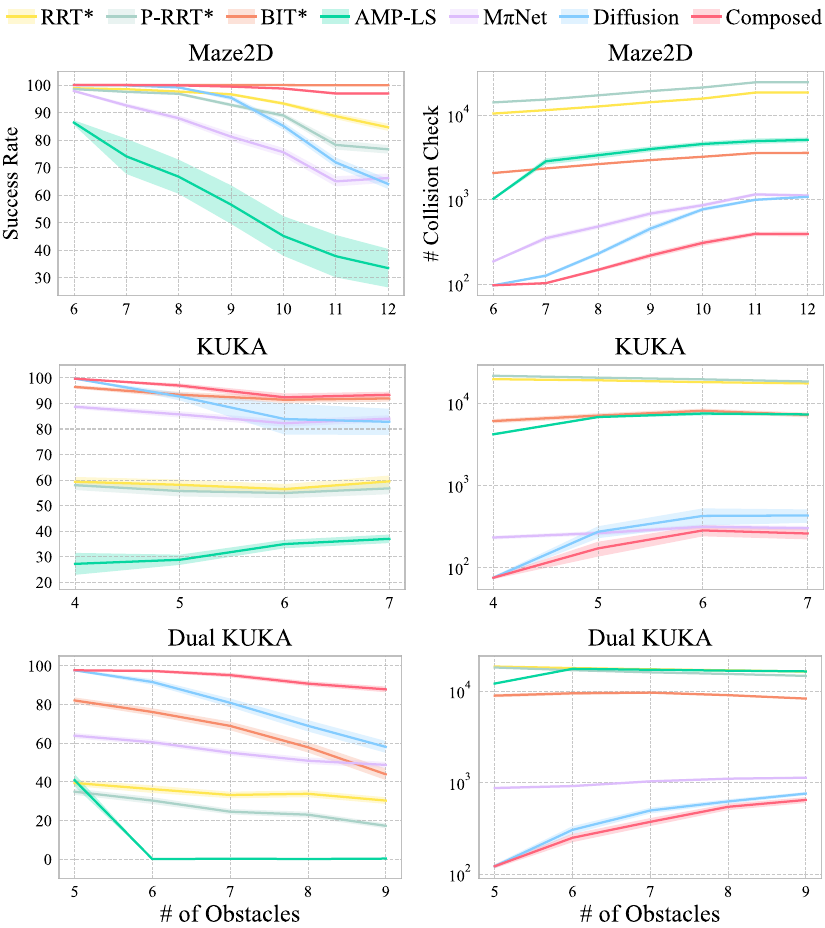}
  \vspace{-14pt}
  \caption{ 
  \textbf{Compositional Generalization.} Quantitative comparisons of different planner on composite environments. The shaded areas represent standard errors. Each graph's leftmost column displays results for environments that contain the same number of obstacles as encountered during training. By composing potentials at test time, our method (red line) can generalize to environments with much more obstacles than training time.
  }
  \label{fig:comp_scatterplot}
  \vspace{-16pt}

\end{figure}

\paragraph{Environments with Concave Obstacles} 
We construct additional Maze2D environments where obstacles are concave to demonstrate the capability of our learned potential motion planner to avoid local minima. As shown in Figure \ref{fig:21-concave7_compare-main-paper}, potential-based method RMP tends to get trapped in local minima in environments filled with concave obstacles. 
In addition, we observe that BIT* is able to find a feasible motion plan, but the morphology of the proposed trajectories are sharp and irregular. 
We further present the quantitative results on base Maze2D environments in Table \ref{tab:concave7-main}, where we include an additional diffusion-based motion planning approach~\citep{carvalho2023motion}.
Notably, our diffusion potential planner is the only learning-based approach that successfully solves every motion planning problems in our evaluation set. 
Although the advanced sampling-based method BIT* can also find all the solutions, its average planning time is significantly longer, taking three times longer than our method in Maze2D -- Concave.
More results are provided in Appendix \ref{sect:app_rm2d_concave}.

\paragraph{Motion Refining} We present quantitative and qualitative results of refining motion plans in Table \ref{tab:replan} and Figure \ref{fig:replan}.
The gain of refining motion plans increases as the dimension of the environment increases. As shown in Table \ref{tab:replan}, the success rate generally increases as we increase the number of refining attempts $R$, but the gain gradually saturates in 10 attempts. In this case, the proposed trajectory probably suffers from a catastrophic collision, and to obtain a successful motion plan, the diffusion potential model might need to resample a trajectory from pure noise.

\subsection{Performance on Composite Environments}\label{subsec:comp}

\paragraph{Composing Obstacles} 
We first evaluate the compositionality by adding obstacles to the environments. 
Qualitative results of composite Maze2D environments are given in Figure \ref{fig:6-comp_rm2d}, where we train our model on 6 obstacles and evaluate on environments with up to 11 obstacles. 
Each orange block represents an extra obstacle added to the test time environments. 
As we can see, the composed model effectively proposes different trajectories according to the presented obstacles by sampling from the composite potential. We report the quantitative performance on three environments with different configuration space dimensions in Figure \ref{fig:comp_scatterplot}. More results are shown in Appendix \ref{sect:app_comp_same}. In addition, we further include a baseline which directly learns demonstrations of different numbers and types of obstacles in a single diffusion potential function in Appendix \ref{sect:app:learn-all-comb}.

\begin{figure}[t]
  \centering
  \includegraphics[width=1.0\linewidth]{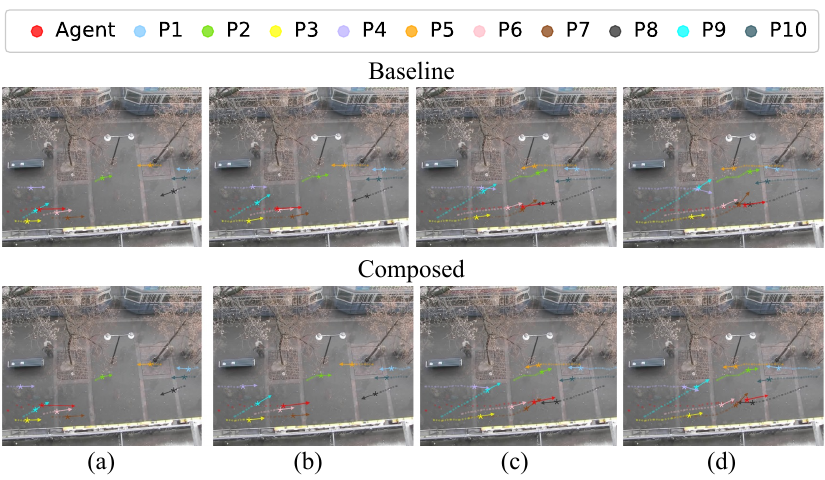}
  \vspace{-15pt}
  \caption{ \textbf{ Qualitative Real World Motion Plans, \textit{Hotel} Scene. } The composed model provides long-horizon motion plan that avoid 10 pedestrians, while only trained on 5 pedestrians. In column (a) and (b), the composed plan is aware of P1 (cyan) and P6 (pink) and overtakes them from above, while the baseline model runs into them. In column (c), the composed motion plan chooses to move faster so as to pass through the intersection with P7 (brown) before P7 arrives, but the baseline motion plan results in a collision due to its slower speed. In column (d), the composed plan choose to go upward to avoid the oncoming P8 (black).
  }
  \label{fig:11-realworld-comp}
  \vspace{-5pt}
\end{figure}

\begin{table}[t] %

\begin{center}
\setlength{\tabcolsep}{2pt}
\centering

\resizebox{\columnwidth}{!}{
\small %
\begin{tabular}{c c c c c c c c c c }
 & \multicolumn{3}{ c }{Base Dynamic} & \multicolumn{3}{ c }{Static 1 + Dynamic} & \multicolumn{3}{ c }{Static 2 + Dynamic} \\
\cmidrule(lr){2-4} \cmidrule(lr){5-7} \cmidrule(lr){8-10}
M & Suc & Time & Col & Suc & Time & Col &  Suc & Time & Col \\
\midrule

SIPP  &  69.9 & 32.2 & 1M+  & 70.4 & 185.5 & 1.7M+  &   74.0 & 98.7 & 1.3M+     \\
Ours & \textbf{99.5}  & \textbf{0.13} & \textbf{168.8} & \textbf{96.6} & \textbf{4.31} & \textbf{828.6} & \textbf{97.5} & \textbf{4.23} & \textbf{646.4} \\ %

\bottomrule
\end{tabular}
}
\vspacef{-5pt}
\captionof{table}{\textbf{Quantitative Results on Base Dynamic and Static + Dynamic on Maze2D.} Static 1 and Static 2 refer to two different static Maze2D environments. Our planner can generalize to environments with both static and dynamic obstacles while only separately trained on static environments or dynamic environments.}
\label{tab:dyn_rm2d}

\end{center}
\vspace{-20pt}

\end{table}

\paragraph{Composing Multiple Constraints} 
We then investigate the compositionality to combine two different diffusion potential functions together, i.e., models trained on completely different environments.
Specifically, we separately train a model on 6 small obstacles (\textit{Static 1}) and a model on 3 large obstacles (\textit{Static 2}) and evaluate the composed model on environments where both small and large obstacles are presented. 
Quantitative and qualitative results are shown in Table \ref{tab:diff_static_rm2d} and Figure \ref{fig:43-comp_rm2d-diff}. Moreover, we compose the models trained on static environments with another model trained on dynamic environments, 
by which the composed model can generalize to environments where both static and dynamic obstacles are presented, namely \textit{Static 1 + Dynamic} and \textit{Static 2 + Dynamic}. 
The corresponding quantitative results are shown in Table \ref{tab:dyn_rm2d}.
The planning time limit for SIPP is set to 60s/300s for base/composite dynamic environments.
Please refer to Figure \ref{fig:44-dyn_comp_rm2d-61} and Figure \ref{fig:7-dyn_comp_rm2d} in Appendix \ref{sect:app_comp_diff} for more qualitative results.

\subsection{Performance on Real World Datasets}\label{subsec:realworld}
Finally, we evaluate the effectiveness of our method on the real world ETH/UCY~\citep{pellegrini2010improving, lerner2007crowds} dataset. 
The dataset we used consists of 6 scenes (\textit{ETH, Hotel, Zara01, Zara02, Students01, Students03}), where each scene contains human trajectories in world-coordinates collected by manual annotation from a bird-eye-view camera.
Our focus is to investigate if our model can propose successful trajectories given the start and goal locations of an agent in a random, cluttered street-level real-world interaction.
Specifically, the planner is trained to predict the trajectory of the agent (highlighted in red), conditioned on the trajectories of 5 other pedestrians. The training data contains 5 scenes and the held-out scene is used for evaluation. 
In Figure \ref{fig:10-realworld-base}, we present the qualitative results where 5 other pedestrians are presented.
We also evaluate on 10 presented pedestrians by composing two potential functions each constrained by 5 pedestrians, as illustrated in Figure \ref{fig:11-realworld-comp}. Details settings and qualitative results are presented in Appendix \ref{sect:app_real_world}.

\vspacef{-10pt}
\section{Discussion}
\vspacef{-3pt}
\textbf{Limitations.}  Our existing formulation of potential based diffusion motion planner has several limitations. First, although our motion trajectory is accurate, it is often suboptimal, e.g., there exists a shorter path from start to goal. This may be addressed by adding an additional potential to reach the goal as soon as possible.
Second, our approach to composing potentials scales linearly with the number of composed models, requiring significantly more computation power with additional models. This can be remedied by having different potential operate on shared features in a network.

\textbf{Conclusion.} In this work, we have introduced a potential based diffusion motion planner.
We first formulate our diffusion potential motion planner and describe its connections and advantages over traditional potential based planner. We illustrate the motion planning performance of our approach in terms of success rate, planning time, and the number of collision checks over motion planning problems with dimension of 2D, 7D, 14D. We further illustrate the compositionality of the approach, enabling generalization to both new objects and new combinations of motion constraints. Finally, we illustrate the potential of our work on real world scenes with multi-agent interaction.

\section*{Acknowledgements}
We acknowledge support from NSF grant 2214177; from AFOSR grant FA9550-22-1-0249; from ONR MURI grant N00014-22-1-2740; from ARO grant W911NF-23-1-0034; and from the Samsung Global Research Outreach Program. Yilun Du is supported by a NSF Graduate Fellowship. 
We thank the anonymous reviewers for their careful review.
Our research was conducted using computational resources at the Center for Computation and Visualization at Brown University.

\section*{Impact Statement}
This paper presents work whose goal is to advance the field of Machine Learning. 
There are many potential societal consequences of our work, which we believe should be similar to a generic machine learning paper.
\redt{For example, if the motion pattern in the dataset is biased, the motion plans generated by our model might have similar biases as the training dataset.
No other issues we feel must be specifically highlighted here.
}

\bibliography{example_paper}
\bibliographystyle{icml2024}

\newpage
\appendix
\onecolumn

\section{Appendix}
In this appendix, we first present our dataset details in Section \ref{sect:app_env_dataset}. Next, we provide implementation details in Section \ref{app:impl}, such as model architecture, training and evaluation setups, hyperparameters.
Further, We show additional quantitative and qualitative results in Section \ref{sect:app_add_results}, including base environments, composite environments, and real-world datasets. \redt{Lastly, in Section \ref{sect:app_cond_indep}, we give a proof on the optimality of composing potentials of sets of constraints.}

\subsection{Dataset Details} \label{sect:app_env_dataset}
In this section, we present details of the three base environments. 
Each dataset consists of feasible trajectories of randomly sampled start and goal states. All datasets are collected by BIT*\citep{gammell2015batch}. All environments except Maze2D are simulated via PyBullet~\citep{coumans2021}. 

\paragraph{Maze2D} The workspace is a 5 $\times$ 5 square and the agent is a point-robot whose state is its x-y coordinate in the workspace. In the base Maze2D environment (\textit{Static 1}), the obstacles are 6 square blocks of size 1 $\times$ 1. We construct another Maze2D environment (\textit{Static 2}) where obstacles are 3 square blocks of size 1.4 $\times$ 1.4. In addition, we construct a Maze2D environment with 7 concave obstacles.
For simplicity, the volume of the robot agent is ignored -- collision happens when the location is inside the region of an obstacle. The training data contains 3,000 different environment configurations and 25,000 states for each environments (approximately 500 trajectories).

\paragraph{KUKA} One KUKA LBR iiwa robotic arm is placed at the center $(0,0,0)$ in world coordinate and obstacles are given as cubic of length 0.40 meter and are randomly placed in the surrounding of the robot. The training data contains 2,000 different environment configurations and 25,000 states for each environments.

\paragraph{Dual KUKA} Two KUKA LBR iiwa robotic arms are placed side by side on a tabletop. One is at world coordinate $(-0.5, 0, 0)$ and the other is at $(0.5, 0, 0)$. Obstacles are given as cubic of length 0.44 meters and are randomly placed in the surrounding of the robots. The training data contains 2,500 different environment configurations and 25,000 states for each environment.

\paragraph{ETH/UCY} The dataset we used contains 6 scenes, where in each scene there are hundreds of street-level pedestrians trajectories.
In each scene, a video is recorded using a fixed bird-eye-view camera facing the street, and the coordinates of each pedestrian are manually annotated according to the recorded video. We use real-world pedestrian trajectories of 5 scenes to train our model and evaluate on the one held-out scene.

\subsection{Implementation details}\label{app:impl}

\paragraph{Software:} The computation platform is installed with Red Hat 7.9, Python 3.8, PyTorch 1.10.1, and Cuda 11.1

\paragraph{Hardware:} For each of our experiments, we used 1 RTX 3090 GPU.

\subsubsection{Energy-based Diffusion Model}
\paragraph{Model Architecture} The diffusion potential model $f_\theta$ consists of a CNN trajectory denoiser based on U-Net similar to \citep{ajay2023is} and a constraint (i.e., environment configuration) encoder. 
The U-Net contains repeated residual blocks where each block consists of two temporal convolutions followed by GroupNorm and SiLU nonlinearity~\citep{hendrycks2016gaussian}.
The constraint encoder uses the Transformer encoder structure~\citep{vaswani2017attention}, whose input is a set of obstacle locations for static environments or obstacle trajectories for dynamic environments. We remove the positional embedding, since the obstacles information should be permutation invariance with each other. We concatenate the learned class token from transformer with the time embedding and feed the concatenated tensor to temporal convolution blocks in U-Net for denoising.  
More details of the models are shown in Table \ref{tab:unet_hyper} and Table \ref{tab:vit_hyper}.
Note that we do not further explore the selection of the concrete model architecture, but we believe that some more advanced architectures could further improve our performance. %

\vspace{10pt}
\begin{center}
\begin{minipage}[t]{0.49\textwidth}
\setlength{\tabcolsep}{4pt}
\centering
    \begin{tabular}{cc}
         Hyperparameters & Value\\
         \midrule
         Base Feature Channels & 64  \\
         Feature Dimension Scale &  (64, 256, 512) \\
         Groups in GroupNorm & 8\\
         Nonlinearity   & SiLU \\
        \bottomrule
    \end{tabular}
    \label{tab:unet_hyper}
\captionof{table}{Hyperparameter of U-Net.}
\end{minipage}
\hfill
\begin{minipage}[t]{0.49\textwidth}
    \setlength{\tabcolsep}{2pt}
    \centering
        \begin{tabular}{cc}
         Hyperparameters & Value\\
         \midrule
         Base Embedding Channel & 64  \\
         Transformer Layers & 3 \\
         Attention Heads & 1 \\
         Nonlinearity   & SiLU \\
        \bottomrule
    \end{tabular}
    \label{tab:vit_hyper}
    \captionof{table}{Hyperparameter of Constraint Encoder.}
\end{minipage}
\end{center}
\vspace{10pt}

\paragraph{Energy Parameterization} To encode the energy $E(\cdot)$ of as in \eqref{eqn:train_obj}, we use L2 energy-parameterization as given in \eqref{eq:energy_l2}. For more details on Energy-based Diffusion Model, please refer to~\citep{du2023reduce}.
\begin{equation}
    \label{eq:energy_l2}
    E_\theta^{L2}(x, t) =  \frac{1}{2} || f_\theta(x, t) ||^2
\end{equation}

\subsubsection{Training Details}
\paragraph{Training Pipeline} In training, the dataloader randomly samples trajectories of length equal to the training horizon from the whole dataset. We provide detailed hyperparameters for training our model in Table \ref{tab:train-hyperparam}. We do not apply any hyperparameter search nor learning rate scheduler. 
The training time of our model is approximately two days, but we observe that the performance is close to saturation within one day.
\begin{table}[h]
    \centering
    \begin{tabular}{cc}
         Hyperparameters & Value\\
         \midrule
         Horizon& 48  \\
         Diffusion Time Step & 100 \\
         Probability of Condition Dropout & 0.2  \\
         Iterations & 2M \\
         Batch Size & 512    \\
         Optimizer  & Adam    \\
         Learning Rate  & 2e-4\\
        \bottomrule
    \end{tabular}
    \caption{Hyperparameters of Diffusion Potential Motion Planner (Training) }
    \label{tab:train-hyperparam}
\end{table}

\subsubsection{Evaluation Details}\label{app:eval-details}
\paragraph{Our Evaluation Pipeline} The input of the planner is the start state, goal state, and the environment constraints (e.g., obstacle locations represented in workspace), and the output is the proposed trajectory. 
In test time, the number of denoising timestep is set to 10 by using DDIM \citep{song2020denoising}. The intermediate noise scale eta in DDIM is set to 0 for base environments and 1 for composite environments.
The evaluation pipeline of our model consists of three phases: Propose Motion Plan Candidates, Candidate Selection, and Motion Refining.
The planner first generates multiple candidate trajectories using Algorithm \ref{alg:comb}. It then accesses the environment and performs collision check to select a successful trajectory from the candidates. Finally, if no desired candidate is found, it will execute the motion refining as in Algorithm \ref{alg:replan}.
Hyperparameters used in evaluation is detailed in Table \ref{tab:eval-hyperparam-rm2d} and Table \ref{tab:eval-hyperparam-kuka}. We observe that our planner can directly solve all the problems in Maze2D, hence we do not use replanning during the evaluation of Maze2D environments.

\begin{table}[ht]
\centering
\begin{minipage}{0.45\linewidth}
\centering
\begin{tabular}{cc}
     Hyperparameters & Value\\
     \midrule
     Horizon &  48 \\
     DDIM Time Step & 8 \\
     DDIM eta & 0.0 \\
     Guidance Scale & 2.0 \\
     \# of Trajectory Candidate & 20 \\
     \# of Refine Attempts $R$ & 0 \\
     Refine Noise Scale $k$ & -- \\
    \bottomrule
\end{tabular}
\caption{
Hyperparameters of Diffusion Potential Motion Planner (Evaluation on Maze2D)
}

\label{tab:eval-hyperparam-rm2d}
\end{minipage}
\hfill %
\begin{minipage}{0.45\linewidth}
\centering
\begin{tabular}{cc} %

     Hyperparameters & Value\\
     \midrule
     Horizon &  52 \\
     DDIM Time Step & 10 \\
     DDIM eta & 0.0 \\
     Guidance Scale & 2.0 \\
     \# of Trajectory Candidate & 20 \\
     \# of Refine Attempts $R$ & 5 \\
     Refine Noise Scale $k$ & 3 \\
    \bottomrule

\end{tabular}
\caption{
Hyperparameters of Diffusion Potential Motion Planner (Evaluation on KUKA and Dual KUKA)
}
\label{tab:eval-hyperparam-kuka}
\end{minipage}
\end{table}

\paragraph{Baselines Evaluation Pipeline} We try our best to re-implement every baseline and follow their original settings. For MPNet~\citep{qureshi2019motion}, we follow their implementation and use bi-directional path generation in test time. As for AMP-LS, we implement a Variational Auto-Encoder (VAE)~\citep{kingma2013auto} to encode the robot pose state and leverage GECO loss \citep{rezende2018taming} in path optimization. M$\pi$Net does not provide a replan scheme in their design. For fair comparison, we boost M$\pi$Net with replan by backtracing to previous timestep and adding a small random noise for restart when any collisions are detected.

\section{Additional Results} \label{sect:app_add_results}
In this section, we provide more quantitative and qualitative motion planning results.  In Section \ref{sect:app_base_env}, we present additional planning results on the base environments. 
In Section \ref{sect:app_comp_same}, we show planning performance on composite same environments. 
In Section \ref{sect:app_comp_diff}, we provide planning performance on composite different environments, including composing two different static models and composing a static model and a dynamic model.
In Section \ref{sect:app_rm2d_concave}, we show motion planning performance on a more challenging Maze2D environment with concave obstacles. 
In Section \ref{sect:app_real_world}, we present more quantitative and qualitative evaluation results on the real-world dataset.

\subsection{Performance on Base Environment}\label{sect:app_base_env}
We provide detailed numerical motion planning results on three base environments in Table \ref{tab:base}. The corresponding bar plot visualization is given in Figure \ref{fig:base_barplot}. Besides, we provide an additional GNN-based motion planning baseline \citep{yu2021reducing} in the table below.
We report the mean and standard error on three main motion planning metrics: success rate, planning time and number of collision checks.
Our method consistently outperforms all other baselines on success rate. Especially, in the hardest Dual KUKA environment, our planner outperforms other learning-based baseline by 40-50\% on success rate, and even outperforms the advance sampling-based method BIT* by 15\% while with one order of magnitude less planning time and collision checks.

\begin{table}[H]

\small\setlength{\tabcolsep}{2.5pt}
\centering
\small %
\begin{tabu}{llllllllll}
 & \multicolumn{3}{ c }{Maze2D} & \multicolumn{3}{ c }{KUKA} & \multicolumn{3}{ c }{Dual KUKA} \\
\cmidrule(lr){2-4} \cmidrule(lr){5-7} \cmidrule(lr){8-10}

Method & Success & Time & Check &  Success & Time & Check & Success & Time & Check \\
\midrule

RRT* &  \numerr{98.8}{0.3} &  \numerr{0.95}{0.02} &  \numerr{10453.4}{156.8} &  \numerr{59.4}{1.9} &  \numerr{3.15}{0.06} &  \numerr{19508.6}{278.6} &  \numerr{39.5}{1.6} &  \numerr{4.16}{0.04} &  \numerr{18560.5}{112.6} \\
P-RRT* &  \numerr{98.5}{0.3} &  \numerr{1.17}{0.02} &  \numerr{14158.9}{225.8} &  \numerr{58.0}{1.9} &  \numerr{3.44}{0.06} &  \numerr{21471.4}{269.1} &  \numerr{35.0}{1.6} &  \numerr{4.54}{0.02} &  \numerr{18117.7}{78.8} \\
BIT* &  \numerr{100.0}{0.0} &  \numerr{0.21}{0.00} &  \numerr{2067.4}{24.4} &  \numerr{96.5}{0.7} &  \numerr{0.86}{0.04} &  \numerr{6047.6}{364.7} &  \numerr{82.1}{1.6} &  \numerr{2.37}{0.08} &  \numerr{8925.4}{365.0} \\
AMP-LS &  \numerr{86.3}{0.9} &  \numerr{1.41}{0.11} &  \numerr{1025.4}{30.2} &  \numerr{27.2}{4.3} &  \numerr{3.89}{0.35} &  \numerr{4176.4}{131.0} &  \numerr{40.9}{3.2} &  \numerr{7.38}{0.72} &  \numerr{12091.3}{230.0} \\
MPNet &  \numerr{85.6}{0.9} &  \numerr{0.41}{0.01} &  \numerr{6315.7}{203.0} &  \numerr{76.0}{3.9} &  \numerr{0.10}{0.02} &  \numerr{885.4}{135.5} &  \numerr{42.0}{4.1} &  \numerr{0.35}{0.04} &  \numerr{1750.2}{195.6} \\
M$\pi$Net &  \numerr{97.9}{0.3} &  \numerr{0.07}{0.00} &  \numerr{178.7}{7.8} &  \numerr{88.7}{0.8} &  \numerr{0.22}{0.01} &  \numerr{232.5}{11.6} &  \numerr{63.9}{1.4} &  \numerr{0.72}{0.02} &  \numerr{875.2}{26.7} \\
GNN & \numerr{100.0}{0.0} & \numerr{0.11}{0.00} & \numerr{1043.3}{5.5} &
\numerr{98.9}{0.4}  & \numerr{0.38}{0.01} & \numerr{2207.4}{153.6} &
\numerr{94.6}{0.6}  & \numerr{1.22}{0.03} & \numerr{6269.1}{344.7} 
\\
\midrule
Ours &  \numerr{100.0}{0.0} &  \numerr{0.12}{0.00} &  \numerr{97.0}{0.3} &  \numerr{99.7}{0.0} &  \numerr{0.13}{0.00} &  \numerr{75.5}{3.1} &  \numerr{97.7}{0.5} &  \numerr{0.27}{0.01} &  \numerr{122.4}{6.1} \\

\bottomrule
\end{tabu}
\caption{\textbf{Quantitative Motion Planning Performance.} Evaluated on 100 unseen environments with 20 motion planning problems in each environment. We report the mean over all environments and the standard error across different environments. The table is the numerical results corresponding to the bar plot visualization in Figure \ref{fig:base_barplot}. 
}
\label{tab:base}
\end{table}

\begin{figure}[H]
  \vspace{-5pt}
  \centering
  \includegraphics[width=0.75\linewidth]{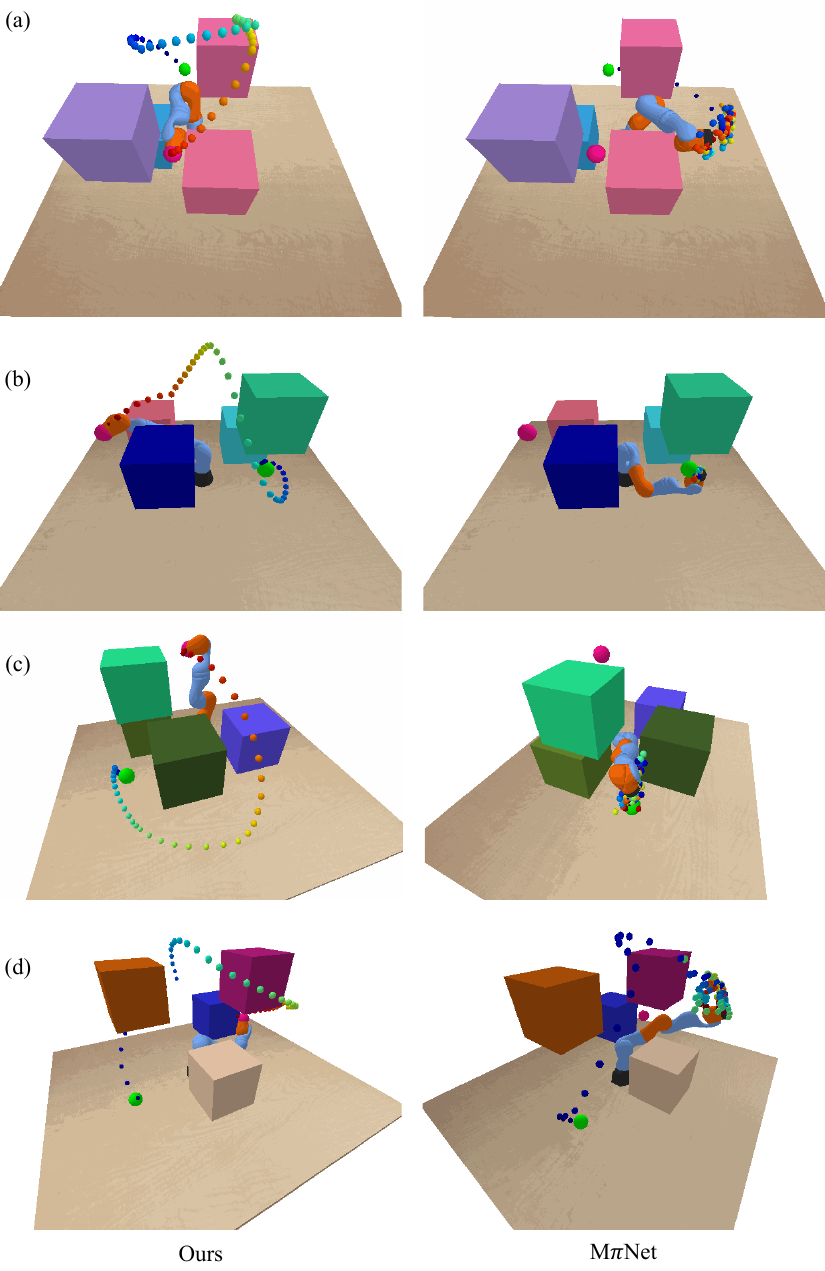}
  \vspace{-7pt}
  \caption{ \textbf{Comparison with M$\pi$Net on KUKA.} Trajectories of 4 motion planning problems.  The large green/pink ball indicates the location of the end effector of the start/goal state. 
  Our planner can generate smooth long-horizon motion trajectories
  and avoid being stuck in local geometry.
  For example, in (a), our planner is able to select a feasible and shorter path passing through the center of the workspace to reach the goal state; and in (c), the trajectory by our planner navigates through narrow passage ways without collision, e.g. travel between the green and purple blocks.
  }
  \label{fig:41-kuka7d-base-app}
\end{figure}

\begin{figure}[H]
  \centering
  \includegraphics[width=0.75\linewidth]{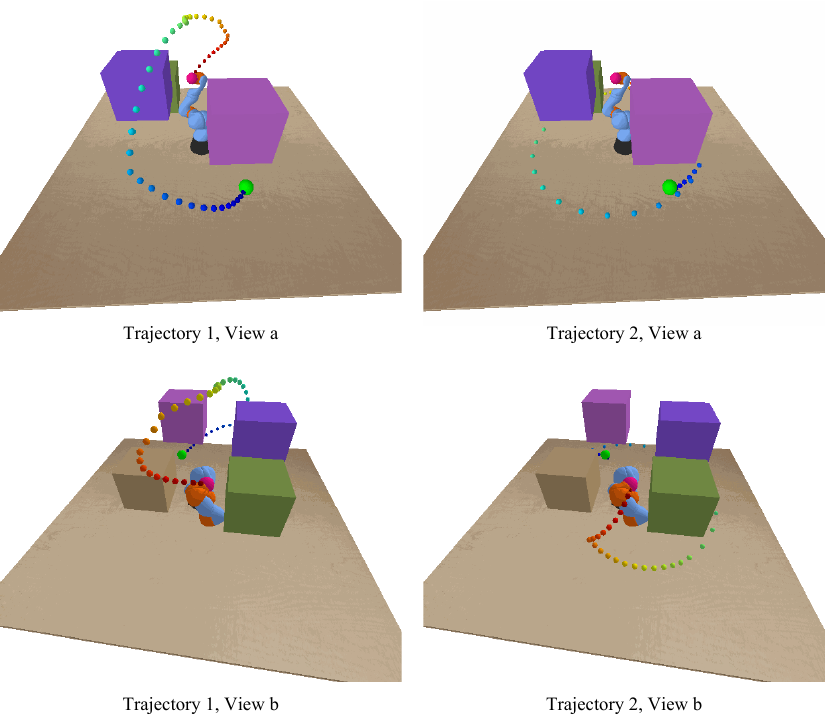}
  \caption{ \textbf{Flexible Trajectory Morphology.} Our method can generate trajectories with various morphological shapes. The large green/pink ball indicates the location of the end effector of the start/goal state. View a and b represent two different viewing angles of the same trajectory. In the figure, Trajectory 1 and Trajectory 2 are generated under the same constraints (e.g., start state, goal state, obstacles locations) but with different starting noise, resulting in different route selection. Trajectory 1 routes through the central narrow passage between the two purple blocks and arrives the pink ball from above. Trajectory 2 avoids the obstacles by going under the blocks and finally arrives the same goal state from below.
  }
  \label{fig:42-kuka7d-morp-app}
\end{figure}

In addition, we present 4 qualitative comparisons with the state-of-the-art learning-based motion planner M$\pi$Net on the KUKA environments in Figure \ref{fig:41-kuka7d-base-app} at the end of this section. 
In Figure \ref{fig:42-kuka7d-morp-app}, we demonstrate the versatile morphology of our motion trajectories on the same motion planning problem.

\subsection{Performance on Composite Same Environment}\label{sect:app_comp_same}
\begin{figure}[h]
  \centering
  \includegraphics[width=1.0\linewidth]{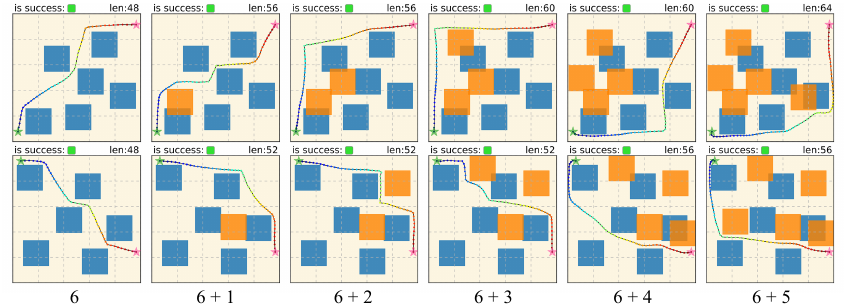}
  \vspace{-10pt}
  \caption{ \textbf{Compositional Generalization over Increasing Obstacles.} The green/pink star indicates the start/goal state. Our planner is trained with a dataset with only 6 obstacles and can generalize to scenarios with more obstacles by directly composing potentials in test time (without any re-training). The generated trajectories demonstrate various morphology and reach the goals with near optimal paths. }
  \label{fig:40-comp_rm2d-app}
\end{figure}

We construct \textit{composite same environments} with more obstacles of the same type than training phase environments. Since more obstacles are presented, these evaluation environments are out of training distribution and hence more challenging.
The locations of each obstacle in the evaluation environments are random and unseen to the model. We further provide a baseline \textit{Diffusion} in which we do not compose potentials.
The quantitative results are shown in Table \ref{tab:comp_table_rm2d_app}, \ref{tab:comp_table_kuka_app}, \ref{tab:comp_table_dual_kuka_app}, and qualitative results are shown in Figure \ref{fig:40-comp_rm2d-app}.
Our composed model demonstrates superior performance over all benchmarks, which is consistent with the expectation that composition enables more effective generalization to more obstacles.

\redtb{
    As presented in Section \ref{sect:diffusion_composition}, when generalizing to more obstacles, we might split the obstacles into several groups before composition. 
    In our experiments, the splitting of obstacles is random, and we do not observe correlations between performance and the ways of splitting. 
    
}

\begin{table}[H]

\small\setlength{\tabcolsep}{5.5pt}
\centering
\small %
\begin{tabular}{c l l l l l l l}
& & \multicolumn{2}{ c }{6 + 1} & \multicolumn{2}{ c }{6 + 2} & \multicolumn{2}{ c }{6 + 3} \\
\cmidrule(lr){3-4} \cmidrule(lr){5-6} \cmidrule(lr){7-8}

 & Method & Success & Check &  Success & Check & Success & Check \\
\midrule

\multirow{7}{*}{Maze2D}

& RRT* &  \numerr{98.5}{0.3} &  \numerr{11467.3}{176.8} &  \numerr{97.7}{0.4} &  \numerr{12675.6}{238.9} &  \numerr{96.7}{0.5} &  \numerr{14247.8}{302.1} \\
& P-RRT* &  \numerr{97.5}{0.3} &  \numerr{15308.9}{257.3} &  \numerr{96.8}{0.5} &  \numerr{17176.1}{324.3} &  \numerr{92.8}{0.7} &  \numerr{19280.9}{429.9} \\
& BIT* &  \numerr{100.0}{0.0} &  \numerr{2340.0}{35.4} &  \numerr{100.0}{0.0} &  \numerr{2635.7}{40.6} &  \numerr{100.0}{0.0} &  \numerr{2942.3}{53.7} \\
& AMP-LS &  \numerr{74.1}{6.4} &  \numerr{2845.2}{247.0} &  \numerr{66.7}{6.2} &  \numerr{3357.6}{298.0} &  \numerr{56.5}{7.0} &  \numerr{3973.5}{290.0} \\
& M$\pi$Net &  \numerr{92.6}{0.8} &  \numerr{349.7}{23.3} &  \numerr{88.0}{1.1} &  \numerr{481.2}{30.1} &  \numerr{81.2}{1.5} &  \numerr{685.3}{45.5} \\
& Diffusion &  \numerr{99.9}{0.1} &  \numerr{126.1}{5.6} &  \numerr{99.2}{0.2} &  \numerr{229.7}{11.8} &  \numerr{95.4}{0.9} &  \numerr{454.8}{34.2} \\
& Composed &  \numerr{100.0}{0.0} &  \numerr{102.8}{2.6} &  \numerr{99.9}{0.1} &  \numerr{147.9}{6.4} &  \numerr{99.5}{0.3} &  \numerr{218.8}{15.1} \\

&\\

& & \multicolumn{2}{ c }{6 + 4} & \multicolumn{2}{ c }{6 + 5} & \multicolumn{2}{ c }{6 + 6} \\
\cmidrule(lr){3-4} \cmidrule(lr){5-6} \cmidrule(lr){7-8}

 & Method & Success & Check &  Success & Check & Success & Check \\
\midrule

\multirow{7}{*}{Maze2D}

& RRT* &  \numerr{93.3}{0.6} &  \numerr{15729.8}{307.4} &  \numerr{88.7}{1.1} &  \numerr{18508.4}{449.0} &  \numerr{84.7}{1.3} &  \numerr{18598.5}{416.9} \\
& P-RRT* &  \numerr{89.0}{0.9} &  \numerr{21280.1}{398.9} &  \numerr{78.2}{1.7} &  \numerr{24471.1}{526.0} &  \numerr{76.7}{1.5} &  \numerr{24491.8}{501.8} \\
& BIT* &  \numerr{100.0}{0.0} &  \numerr{3220.2}{59.7} &  \numerr{100.0}{0.1} &  \numerr{3563.4}{79.2} &  \numerr{100.0}{0.1} &  \numerr{3580.7}{61.1} \\
& AMP-LS &  \numerr{45.1}{7.2} &  \numerr{4560.4}{310.0} &  \numerr{37.8}{7.7} &  \numerr{4928.6}{361.0} &  \numerr{33.5}{7.0} &  \numerr{5107.2}{341.0} \\
& M$\pi$Net &  \numerr{75.6}{1.5} &  \numerr{860.7}{47.3} &  \numerr{65.0}{1.8} &  \numerr{1154.3}{50.8} &  \numerr{66.2}{1.6} &  \numerr{1120.9}{48.0} \\
& Diffusion &  \numerr{85.2}{1.4} &  \numerr{769.7}{41.4} &  \numerr{71.9}{1.8} &  \numerr{997.0}{38.1} &  \numerr{64.0}{1.8} &  \numerr{1080.1}{31.9} \\
& Composed &  \numerr{98.8}{0.3} &  \numerr{308.2}{22.1} &  \numerr{97.0}{0.5} &  \numerr{393.9}{23.0} &  \numerr{97.0}{0.5} &  \numerr{392.7}{25.7} \\

\bottomrule

\end{tabular}
\caption{
\textbf{Compositional Generalization over Increased Obstacles in Maze2D.}  In the top row, the \textit{left digit} represents the number of obstacles that the model is trained on; the \textit{right digit} denotes the number of extra obstacles added to the evaluation environments.
Unlike other learning-based planners, whose performance significantly declines as the number of obstacles increases, our method sustains a near-optimal success rate even when the quantity of obstacles doubles.
}

\label{tab:comp_table_rm2d_app}
\end{table}

\begin{table}[H]

\small\setlength{\tabcolsep}{5.5pt}
\centering
\small %
\begin{tabular}{c l l l l l l l}
& & \multicolumn{2}{ c }{4 + 1} & \multicolumn{2}{ c }{4 + 2} & \multicolumn{2}{ c }{4 + 3} \\
\cmidrule(lr){3-4} \cmidrule(lr){5-6} \cmidrule(lr){7-8}

 & Method & Success & Check &  Success & Check & Success & Check \\
\midrule

\multirow{7}{*}{KUKA} 

& RRT* &  \numerr{58.1}{2.1} &  \numerr{18920.7}{281.3} &  \numerr{56.5}{2.0} &  \numerr{17964.2}{230.7} &  \numerr{59.5}{2.3} &  \numerr{17339.7}{261.3} \\
& P-RRT* &  \numerr{55.7}{2.1} &  \numerr{20356.2}{254.2} &  \numerr{55.0}{2.1} &  \numerr{19420.8}{216.0} &  \numerr{56.8}{2.3} &  \numerr{18287.8}{203.3} \\
& BIT* &  \numerr{93.4}{1.1} &  \numerr{7045.6}{483.9} &  \numerr{91.5}{1.2} &  \numerr{8055.9}{505.9} &  \numerr{92.0}{1.2} &  \numerr{7176.8}{477.9} \\
& AMP-LS &  \numerr{28.8}{2.0} &  \numerr{6767.0}{154.0} &  \numerr{35.0}{1.6} &  \numerr{7461.5}{401.0} &  \numerr{37.0}{1.7} &  \numerr{7277.4}{150.0} \\
& M$\pi$Net &  \numerr{85.7}{1.2} &  \numerr{262.6}{16.0} &  \numerr{82.2}{1.4} &  \numerr{314.7}{19.3} &  \numerr{84.0}{1.3} &  \numerr{301.7}{18.6} \\
& Diffusion &  \numerr{92.7}{1.5} &  \numerr{273.3}{45.1} &  \numerr{83.9}{6.1} &  \numerr{425.5}{99.6} &  \numerr{82.8}{5.1} &  \numerr{430.8}{79.8} \\
& Composed &  \numerr{97.0}{0.8} &  \numerr{171.5}{35.3} &  \numerr{92.4}{1.5} &  \numerr{283.4}{43.2} &  \numerr{93.4}{1.2} &  \numerr{259.9}{39.4} \\
\bottomrule

\end{tabular}
\caption{
\textbf{Compositional Generalization over Increased Obstacles in KUKA.}
In the top row, the \textit{left digit} represents the number of obstacles that the model is trained on; the \textit{right digit} denotes the number of additional obstacles. By composing potentials, our planner surpasses all the baselines by a margin.
}
\label{tab:comp_table_kuka_app}
\end{table}

\begin{table}[H]

\small\setlength{\tabcolsep}{3pt}
\centering
\small %
\begin{tabular}{c l l l l l l l l l}
& & \multicolumn{2}{ c }{5 + 1} & \multicolumn{2}{ c }{5 + 2} & \multicolumn{2}{ c }{5 + 3}  & \multicolumn{2}{ c }{5 + 4} \\
\cmidrule(lr){3-4} \cmidrule(lr){5-6} \cmidrule(lr){7-8} \cmidrule(lr){9-10}

 & Method & Success & Check &  Success & Check & Success & Check & Success & Check \\
\midrule

\multirow{7}{*}{\makecell{Dual \\ KUKA}}

& RRT* &  \numerr{36.2}{1.9} &  \numerr{17843.8}{110.8} &  \numerr{33.2}{1.7} &  \numerr{17311.8}{93.9} &  \numerr{33.9}{1.8} &  \numerr{16847.9}{81.1} &  \numerr{30.4}{1.7} &  \numerr{16355.9}{72.6} \\
& P-RRT* &  \numerr{30.4}{1.8} &  \numerr{16921.2}{75.2} &  \numerr{24.6}{1.4} &  \numerr{16029.4}{74.3} &  \numerr{23.1}{1.5} &  \numerr{15387.9}{69.0} &  \numerr{17.3}{1.3} &  \numerr{14667.8}{81.9} \\
& BIT* &  \numerr{76.2}{1.9} &  \numerr{9470.5}{376.1} &  \numerr{69.0}{2.2} &  \numerr{9631.7}{284.6} &  \numerr{57.9}{2.7} &  \numerr{9034.9}{256.0} &  \numerr{44.0}{3.0} &  \numerr{8290.1}{176.1} \\
& AMP-LS &  \numerr{0.1}{0.0} &  \numerr{17512.0}{180.0} &  \numerr{0.3}{0.1} &  \numerr{17107.4}{219.0} &  \numerr{0.1}{0.1} &  \numerr{16725.1}{233.0} &  \numerr{0.3}{0.1} &  \numerr{16419.1}{210.0} \\
& M$\pi$Net &  \numerr{60.5}{1.4} &  \numerr{919.4}{27.5} &  \numerr{55.1}{1.5} &  \numerr{1034.9}{27.6} &  \numerr{51.0}{1.7} &  \numerr{1106.4}{30.3} &  \numerr{48.8}{1.5} &  \numerr{1134.0}{27.2} \\
& Diffusion &  \numerr{91.7}{1.5} &  \numerr{305.6}{29.8} &  \numerr{80.9}{2.2} &  \numerr{497.1}{35.5} &  \numerr{69.0}{2.9} &  \numerr{624.0}{35.8} &  \numerr{58.1}{3.0} &  \numerr{761.0}{31.0} \\
& Composed &  \numerr{97.3}{0.7} &  \numerr{250.4}{26.2} &  \numerr{95.2}{1.1} &  \numerr{373.9}{37.1} &  \numerr{90.8}{1.3} &  \numerr{545.7}{43.7} &  \numerr{87.8}{1.5} &  \numerr{648.4}{47.2} \\

\bottomrule

\end{tabular}
\caption{
\textbf{Compositional Generalization over Increased Obstacles in Dual KUKA.}
In the top row, the \textit{left digit} represents the number of obstacles that the model trained on; the \textit{right digit} denotes the number of additional obstacles. In this most difficult Dual KUKA environment, our planner outperforms other baselines by a larger margin. From 5 + 1 to 5 + 4, the advanced sampling-based method BIT* drops by 32\%, with a success rate of 44\%, while our method only drops by less than 10\%, achieving 87.8\% in success rate.
}
\label{tab:comp_table_dual_kuka_app}
\end{table}

\subsection{Performance on Composite Different Environment}\label{sect:app_comp_diff}
We present extra experiment results on composing two different models. 
Each model is separately trained on one single obstacle type and we evaluate them on more complex environments with multiple obstacle types. 

\begin{figure}[H]
  \centering
  \includegraphics[width=0.85\linewidth]{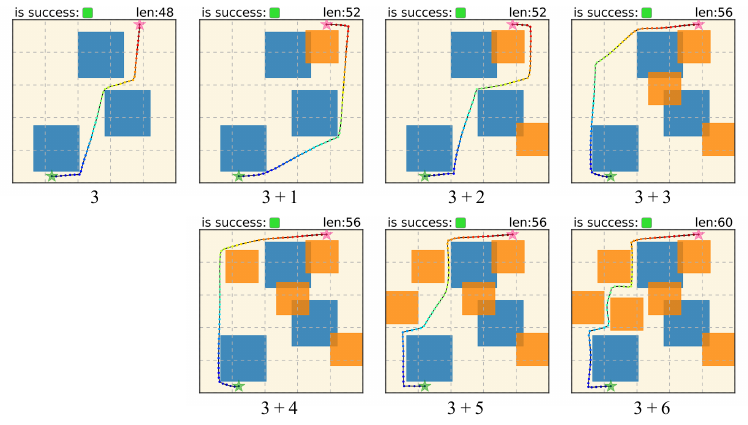}
  \caption{ \textbf{Compositional Generalization over Different Obstacles.} The green/pink star indicates the start/goal state. Two separately trained models are composed: a model \textit{only} trained on large blocks (blue) of size $1.4 \times 1.4$ and a model \textit{only} trained on smaller blocks (orange) $1 \times 1$. By composing two models, our planner can generalize to more complex scenarios with various obstacles in test time (without any re-training). The generated trajectories demonstrate various morphology and reach the goal with smooth and short paths. }
  \label{fig:43-comp_rm2d-diff}
\end{figure}

\paragraph{Static 1 + Static 2} In Table \ref{tab:diff_static_rm2d},  we present quantitative generalization results of composing two different static Maze2D models. Static 1 is a model only trained on obstacles of size $1 \times 1$ and Static 2 is a model only trained on obstacles of size $1.4 \times 1.4$. The composite environment \textit{Static 1 + Static 2} contains six $1 \times 1$ obstacles and three $1.4 \times 1.4$.
As the setting in the evaluation of the base environments, we evaluate on 100 different environments with 20 randomly generated start and goal in each environment. Obstacles in all environments are randomly placed.  No extra training is required. We also present qualitative results over increasing obstacles in the composite different environment (Static 1 + Static 2) in Figure \ref{fig:43-comp_rm2d-diff}.

\begin{table}[h]
    \setlength{\tabcolsep}{7pt}
    \centering
    \begin{tabular}{c c c }
     & \multicolumn{2}{ c }{Static 1 + Static 2 } \\
    \cmidrule(lr){2-3}
    Method & Success & Check \\
    \midrule
    RRT* & 90.3  & 18495.1 \\
    P-RRT* & 82.8   & 24959.9 \\
    BIT* & 100.0  & 3486.9 \\
    \midrule
    Ours & 98.9  & 304.7 \\
    \bottomrule
    \end{tabular}
    \captionof{table}{\textbf{Quantitative Results on Maze2D Composite Different Environments.} Static 1 is a model only trained on obstacles of size $1 \times 1$ and Static 2 is a model only trained on obstacles of size $1.4 \times 1.4$. The testing environments contain six $1 \times 1$ obstacles and three  $1.4 \times 1.4$ obstacles. Though trained on different environments separately, the composed model reaches near optimal success rate while requiring one order of magnitude less collision checks than BIT*.}
    \label{tab:diff_static_rm2d}
\end{table}

\begin{figure}[H]
  \centering
  \includegraphics[width=0.85\linewidth]{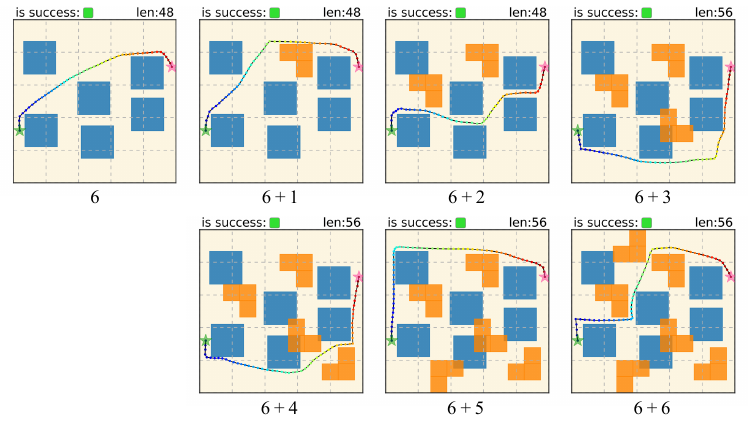}
  \caption{ \textbf{Compositional Generalization over Square and Concave Obstacles.} The green/pink star indicates the start/goal state. Two separately trained models are composed: a model \textit{only} trained on square blocks (blue) of size $1.0 \times 1.0$ and a model \textit{only} trained on concave blocks (orange). Our planner can adaptively propose motion trajectories according to the environments by composing the potentials of square obstacles and concave obstacles.}
  \label{fig:45-comp_rm2d-diff-concave}
\end{figure}

\paragraph{Static 1 + Concave} In Figure \ref{fig:45-comp_rm2d-diff-concave}, we provide qualitative results on composite environments of square obstacles and concave obstacles. In the leftmost column, we show a generated motion trajectory on a base environment with 6 square obstacles only. Following this, we gradually add concave obstacles to the base environment, from 6 + 1 to 6 + 6. Our planner can plan smooth and coherent trajectories accordingly by composing the corresponding potentials.

\begin{figure}[h]
  \centering
  \includegraphics[width=1.0\linewidth]{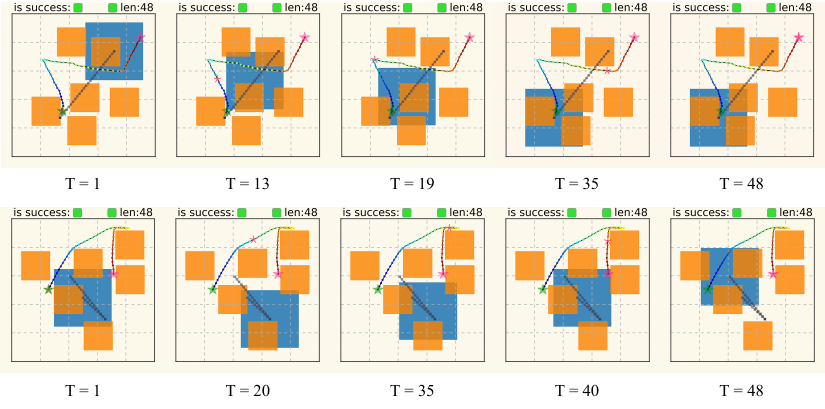}
  \vspace{-20pt}
  \caption{ \textbf{Qualitative Compositionality Generalization over Static 1 + Dynamic Obstacles.} The current position of the agent is shown in the pink asterisk. The grey line indicates the moving trajectory of the dynamic obstacle (the large blue block). In the first row, the planned trajectory goes further left in order to avoid the moving obstacle from the upper right. At T = 19, the trajectory is traveling right while can still avoid the other orange static obstacles. In the second row, the dynamic obstacle is first moving toward the bottom right corner and then going back to the upper left. At around T = 35, the trajectory takes several extra moves, waiting for the blue block passing through the goal position.}
  \label{fig:44-dyn_comp_rm2d-61}
  \vspace{-10pt}
\end{figure}

\paragraph{Static 1 + Dynamic Environment} We evaluate our method on composite different environments which contain both static and dynamic obstacles. Note that our method is only trained on environments that purely consist of static obstacles or dynamic obstacles.
\textit{Static 1} is a model that is only trained on six $1 \times 1$ static obstacles and \textit{Dynamic} is a model that only trained on one moving $2 \times 2$ obstacle. The composite environments contain six $1 \times 1$ static obstacles together with one moving $2 \times 2$ obstacle. The quantitative results are shown in Table \ref{tab:dyn_rm2d} and two qualitative results are shown in Figure \ref{fig:44-dyn_comp_rm2d-61}.

\begin{figure}[H]
  \centering
  \includegraphics[width=1.0\linewidth]{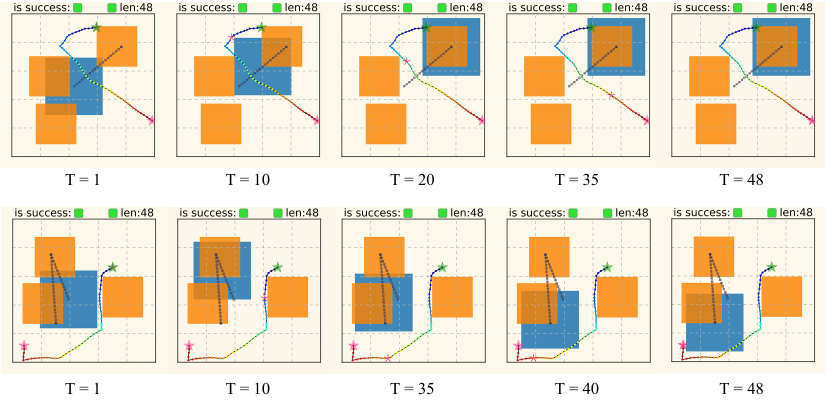}
  \vspace{-20pt}
  \caption{ \textbf{Qualitative Compositionality Generalization over Static 2 + Dynamic Obstacles.} The current position of the agent is shown in the pink asterisk. The grey line indicates the moving trajectory of the dynamic obstacle (the large blue block). Two motion plans are shown. In the first row, the planned trajectory veers further left to circumvent the approaching moving obstacle while still being aware of the other orange static obstacles. In the second row, the dynamic obstacle first moves upwards and then goes downwards. Both phrases of the obstacle's motion can potential block the agent's path, especially when it is going downwards. The planned trajectory first navigates through the gap between the blue block and the orange block, and then reserves enough space and travel near the bottom of the environment to prevent any collisions.
  }
  \label{fig:7-dyn_comp_rm2d}
\end{figure}

\paragraph{Static 2 + Dynamic Environment} 
\textit{Static 2} is a model that is only trained on $1.4 \times 1.4$ static obstacles and \textit{Dynamic} is a model that only trained on one large moving $2 \times 2$ obstacle. The composite environments contain three $1.4 \times 1.4$ static obstacles and one moving $2 \times 2$ obstacle. The quantitative results are shown in Table \ref{tab:dyn_rm2d} and two qualitative results are shown in Figure \ref{fig:7-dyn_comp_rm2d}.

\subsection{Performance on Concave Obstacles}\label{sect:app_rm2d_concave}

We construct Maze2D environments with 7 concave obstacles and have shown the qualitative and quantitative results in Figure \ref{fig:21-concave7_compare-main-paper} and Table \ref{tab:concave7-main} in the main paper.

\begin{figure}[H]
  \centering
  \includegraphics[width=1.0\linewidth]{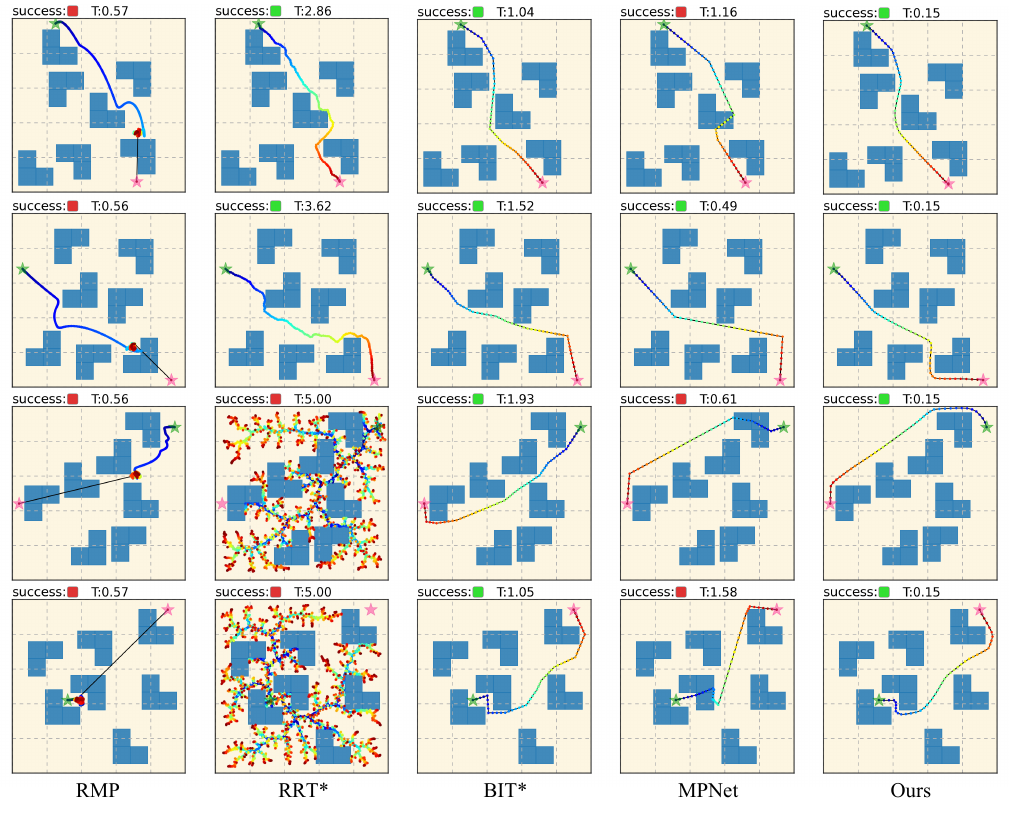}
  \caption{ %
  \textbf{Qualitative Performance on Environments with Concave Obstacles.} 
  The green star indicates the start pose and the red star indicates the goal pose.
  RMP tends to stuck in local minima and fails to reach the goals; 
  RRT* is slow in speed and may reach the planning time limit in some difficult problems; 
  trajectories by MPNet may occasionally go through obstacles in the environments;
  our method is able to generate smooth and low-cost trajectories without collision while being up to 10 times faster than BIT*.
  }
  \label{fig:21-concave7_compare-app}
\end{figure}

Specficially, in Table \ref{tab:concave7-main}, we show the motion planning performance on Convex and Concave obstacles side by side, where RMP~\cite{ratliff2018riemannian} is a traditional potential-based motion planner and MPD is a recent diffusion-based motion planner~\cite{carvalho2023motion}.
Similar to other experiments, all obstacles are randomly placed and no post-training/fine-tuning is required.
We can see that all methods subject to a certain decline in environments with concave obstacles. Notably, our learned diffusion potential motion planner can still solve all the problems, surpassing the pure potential-based method by a margin while requiring significantly less planning time than the state-of-the-art sampling-based planners BIT*. 
In Figure \ref{fig:21-concave7_compare-app}, we present more qualitative results of RMP, RRT*, BIT*, MPNet, and our method on environments with concave obstacles.

\subsection{Performance on Real-World Dataset}\label{sect:app_real_world}
Current robotic path planning problem is usually limited to simulation environments and there is no widely-used real-world benchmark.
Since pedestrian trajectory is naturally a kind of demonstrations of human planning, we leverage the ETH/UCY Dataset to evaluate real-world 
motion planning capability.
Specifically, the entire dataset is comprised of 6 scenes: \textit{ETH, Hotel, Zara01, Zara02, Students01, Students03}, where the models are trained on 5 scenes and tested on the held-out scene.
Similar to our simulation environments, the model takes as input the position of start, goal, and other pedestrians, and outputs a predicted motion plan. 
The predicted motion plan is desired to be as close as possible to the real human trajectory and thus we report the Average Displacement Error (ADE) as defined in \eqref{eq:ade}:
\begin{equation}
\label{eq:ade}
\text{ADE} =  \frac{\sum\limits_{i \in N} \sum\limits_{t \in T} || q_t^i - \hat{q}_t^i ||_2 }{ N \times T }
\end{equation}
where $\hat{q}_t^i \in \mathbb{R}^2$ is the step $t$ in the $i$th predicted path and $q_t^i$ is the corresponding ground truth. ADE measure the similarity between the predicted trajectories and human trajectories. A smaller ADE value indicates the predictions are closer to the real human behavior.

\begin{table}[!h]
    \centering
    \small\setlength{\tabcolsep}{5.3pt}
    \begin{tabular}{c c c c c c c c c c c c c }
         & \multicolumn{2}{ c }{ \textit{ETH} } & \multicolumn{2}{ c }{ \textit{Hotel} } &  \multicolumn{2}{ c }{ \textit{Zara01} } 
        &  \multicolumn{2}{ c }{ \textit{Zara02} } 
        &  \multicolumn{2}{ c }{ \textit{Students01} } 
        &  \multicolumn{2}{ c }{ \textit{Students03} } 
        \\
        \cmidrule(lr){2-3}  \cmidrule(lr){4-5} \cmidrule(lr){6-7} \cmidrule(lr){8-9} \cmidrule(lr){10-11} \cmidrule(lr){12-13}
         Method & ADE $\downarrow$ & Time  & ADE & Time  & ADE & Time  & ADE & Time  & ADE & Time  & ADE & Time \\
         \midrule
         {MPNet} & 18.11  & 0.12 & 28.60 & 0.11
         & 11.06 & 0.12  & 17.22 & 0.11
         & 10.37 & 0.12  & 8.93 & 0.11 \\
        \textcolor{black}{M$\pi$Net}
        & 37.70 & 0.26 & 44.49 & 0.29
        & 1.14 & 0.22 & 13.66  & 0.23 
        & 12.76 & 0.18 & 1.54 & 0.22 \\
        \textcolor{black}{Ours}
        & \textbf{0.94} & \textbf{0.17} & \textbf{5.20} & \textbf{0.17} 
        &\textbf{0.35}& \textbf{0.17} & \textbf{0.38} & \textbf{0.17} 
        & \textbf{0.52} & \textbf{0.17} & \textbf{0.89} & \textbf{0.17}  \\
        \bottomrule
    \end{tabular}
    \caption{ %
    \textbf{Quantitative Results on real-world ETH/UCY Dataset.} 
    We adopt the ADE metric to show the similarity between the predicted motion plans and real human motion trajectories on unseen scenarios. All motion plans are in the world coordinate as given in the dataset.
    Our method can mimic human motion behaviors more precisely in most scenes, while both MPNet and M$\pi$Net cause drastic error compared to real human trajectories.}
    \label{tab:ethucy_ade}
    
\end{table}
The per-scene quantitative performance is shown in Table \ref{tab:ethucy_ade}.
Each scene is captured at different time of a day or different location, resulting in different data distribution. 
The ADE is relatively smaller in \textit{Zara} and \textit{Students} because the model can see a similar counterpart scene at training, e.g., train set includes \textit{Zara01} when tested on \textit{Zara02}.
By contrast, \textit{ETH} or \textit{Hotel} are more unique to other scenes (e.g., contain different scene layout and human behavior patterns) and thus causing higher evaluation error.
As shown in Table \ref{tab:ethucy_ade}, MPNet consistently falls short of the ADE, though with slightly faster speed. We observe that M$\pi$Net can produce reasonable motion plans in many cases, but it occasionally predicts random values, which causes significant deviation from the target and leads to the notbaly larger ADE. Our method demonstrates better generalizability and stability by precisely mimicing the human trajectories in most held-out scenes. We also notice that in \textit{Hotel}, all methods suffer from a severe surge in ADE. Our speculation is that the \textit{Hotel} has different walkable world-coordination and in addition, it contains various unseen types of pedestrians motion pattern, such as slowly pacing people that are chatting or waiting, people stepping on or off the train.

\begin{figure}[H]
  \centering
  \includegraphics[width=1.0\linewidth]{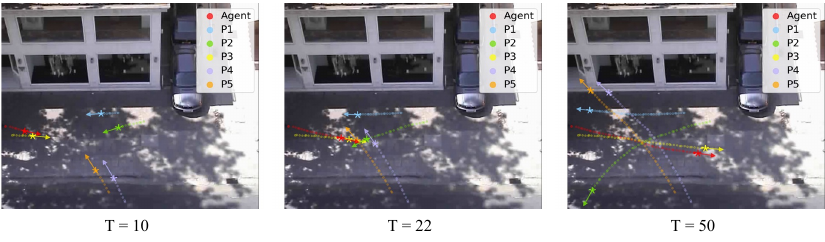}
  \vspace{-10pt}
  \caption{ \textbf{Qualitative Real World Motion Plan, \textit{Zara02} Scene.} Each star represents a pedestrian on the street. Red indicates the trajectory planned by our model, while other colors represent 5 pedestrians in the surrounding. Our motion plan smoothly passes through the intersection without any collision or discontinuity.}
  \label{fig:10-realworld-base} %
\end{figure}

\paragraph{Base Real World Motion Planning} We visualize the motion trajectory planned by our model in \textit{Zara02} scene where five other pedestrians are presented in Figure \ref{fig:10-realworld-base}.  The planner is trained on the other 5 scenes and evaluated on this held-out scene. 
In the given scenario, our agent (highlighted in red) is heading towards the bottom right corner where P2 (yellow) is on its way. Also, the agent is about to enter an intersection with multiple oncoming pedestrians.
In this complex dynamic scene, the trajectory planned by our model first chooses to follow P2 (shown in T = 10) and adjust its pace to let other pedestrians pass through first. 
As shown in in T = 22, our agent successfully crosses the intersection without interrupting any other pedestrians.

\subsection{Comparison to Training on Multiple Types and Numbers of Obstacles}\label{sect:app:learn-all-comb}

\redtb{

In this section, we train an additional baseline which learns all data with different numbers and types of obstacles in a single diffusion potential function. 
Thus, training this baseline requires additional motion planning demonstrations of various obstacle combinations. 
Note that for our proposed compositional motion planning method, models are only trained on a limited number or type of obstacles, while can effectively generalize to various combinations through composition.

We use Maze2D environment as the testbed. 
Let $s$ denote obstacle of type 1 and $l$ denote obstacle of type 2, so that $s6$ indicates 6 obstacles of type 1. We keep all the training and evaluation setups the same as in previous experiments in \ref{sect:app_comp_same}. In Table \ref{tab:app-direct-train-rm2d}, \textit{Direct Training} refers to a model directly trained on various numbers/types of obstacles, and \textit{Composed} refers to the model trained on a fixed number of obstacles and generalizes to novel obstacle combinations via composing potentials in test-time. For further comparison, we include another baseline \textit{Process All Obstacles} in the table, which is also trained on a fixed number of obstacles, but is directly passed in different numbers of obstacles at test-time (This baseline is equivalent to \textit{Diffusion} shown in Table \ref{tab:comp_table_rm2d_app}).

Shown in Table \ref{tab:app-direct-train-rm2d}, we observe that composing models obtains a comparable performance to the \textit{Direct Training} and substantially outperforms \textit{Process All Obstacles}. Note that the training and evaluating distribution is similar for Direct Training, while very different for \textit{Composed}, for example, the training data in \textit{Direct Training} includes the scenarios with 12 obstacles, while \textit{Composed} is only trained on 6 obstacles, highlighting the effectiveness of composition for generalization.

\begin{table}[H]
\small\setlength{\tabcolsep}{4pt} %
\centering
\small %

\begin{tabular}{l l l l l l l l    l l l     }

   & \multicolumn{2}{ c }{  $s6$  } & \multicolumn{2}{ c }{  $s7$  } & \multicolumn{2}{ c }{  $s8$  } & \multicolumn{2}{ c }{  $s9$  } & \multicolumn{2}{ c }{$s10$} \\
\cmidrule(lr){2-3} \cmidrule(lr){4-5} \cmidrule(lr){6-7} \cmidrule(lr){8-9}  \cmidrule(lr){10-11}
Method & Success & Check &  Success & Check & Success & Check  & Success & Check  & Success & Check \\
\midrule
Process All Obstacles  &  100.0 &  97.0 &  99.9 &  126.1 &  99.2 &  229.7 &  95.4 &  454.8 &  85.2 &  769.7 \\
Direct Training  &  100.0 &  99.5 &  100.0 &  86.3 &  100.0 &  95.7 &  100.0 &  118.4 &  99.9 &  140.5  \\
Composed  &  100.0 &  97.0 &  100.0 &  102.8 &  99.9 &  147.9 &  99.5 &  218.8 &  98.8 &  308.2 \\
\bottomrule
\end{tabular}

\vspace{8pt}

\begin{tabular}{l l l l    l l l   l l l l }
   & \multicolumn{2}{ c }{  $s11$  } & \multicolumn{2}{ c }{  $s12$   } & \multicolumn{2}{ c }{  $s6+l1$  } & \multicolumn{2}{c}{  $s6+l2$  } & \multicolumn{2}{ c }{  $s6+l3$   } \\

\cmidrule(lr){2-3} \cmidrule(lr){4-5} \cmidrule(lr){6-7} \cmidrule(lr){8-9}  \cmidrule(lr){10-11}
Method & Success & Check &  Success & Check & Success & Check  & Success & Check  & Success & Check \\
\midrule
Process All Obstacles   &  71.9 &  997.0 &  64.0 &  1080.1  & \multicolumn{2}{ c }{--} & \multicolumn{2}{ c }{--} & \multicolumn{2}{ c }{--}  \\
Direct Training &  98.8 &  227.4 &  98.2 &  264.1
 &  100.0 &  90.4 &  99.9 &  133.8 &  99.8 &  172.4 \\
Composed  &  97.0 &  393.9 &  97.0 &  392.7
 &  99.9 &  184.2 &  99.2 &  304.1 &  98.9 &  304.7 \\
\bottomrule
\end{tabular}

\caption{ %
\textbf{Quantitative Comparison with Model Directly Trained on various Types/Number of Obstacles.}
Motion planning performance on the Maze2D environments with combinations of different types and numbers of obstacles, e.g., $s6 + l2$ denotes an environment with six obstacles of type 1 and two obstacles of type 2. In columns with --, \textit{Processing All Obstacles} baseline is not applicable as different models are composed. Note that in \textit{Direct Training}, the model is trained on all 10 obstacle combinations shown above; while for \textit{Composed}, we only train a model on environments with six obstacles of type 1 and a model on environments with three obstacles of type 2. 
}
\label{tab:app-direct-train-rm2d}
\end{table}

}

\section{Proof of Conditional Independence for Composing Potentials}\label{sect:app_cond_indep}
In Section \ref{sect:diffusion_composition}, we present a way to generalize to multiple unseen out-of-training-distribution constraints by composing corresponding potentials. 
Specifically, in Equation \ref{eqn:diffusion_opt_comb}, we assume that
constraint $C_1$ and $C_2$ are conditional independent. In this section, we will show how our assumption holds in a general case, and thus the compositionality of our planner can be achieved as in \citep{liu2022compositional}.

Assume that two set of constraints are given, $C_1 = \{o_1, o_2, o_3, o_4\}$ and $C_2 = \{ o_3, o_4, o_5, o_6 \}$.
Let $f_{C_i}(q_{1:T})$ denote a probability density function over trajectories, where positive likelihood is uniformly assigned to the trajectory $q_{1:T}$ if it satisfies the constraint $C_i$; otherwise, the likelihood is set to 0. 
Let $\mathcal{J}_{c_i}$ denote the set of trajectories that satisfies constraint $C_i$. Then, we have
\begin{equation}
    f_{C_i}(q_{1:T}) = \begin{cases} 
    \rho_i & \text{if } q_{1:T} \in \mathcal{J}_{c_i} \\ 
    0 & \text{if } q_{1:T} \notin \mathcal{J}_{c_i}
    \end{cases}
\end{equation}
where $\rho_i$ is a small constant. Similarly, we can define $f_{C_1 \cup C_2}$ as the probability density function of trajectories that satisfy both $C_1$ and $C_2$. Clearly, given any trajectory $q_{1:T}$, the probability density of $q_{1:T}$ is positive if and only if both $f_{C_1}(q_{1:T})$ and $f_{C_2}(q_{1:T})$ are positive. More specifically, we have
\begin{equation}
    f_{C_1 \cup C_2}(q_{1:T}) = \gamma f_{C_1}({q_{1:T}}) f_{C_2}({q_{1:T}})
\end{equation}

where $\gamma$ is a constant \redt{(note the proportionality constant in the previous equation).} We can see that the joint probability density function equals to the scaled product of $f_{C_1}$ and $f_{C_2}$. While the above equation doesn't indicate independence, sampling using the score function for the left side of the equation is the same as sampling from the summed score function for the right side of the equation, because the score function is the gradient of the log probability and is invariant to the constant multiplier. 
Therefore, the constant $\gamma$ here will not affect the test-time sampling process 
and the proposed procedure of composing potentials for generalization can be achieved in theory.

\typeout{get arXiv to do 4 passes: Label(s) may have changed. Rerun}

\end{document}